\documentclass[a4paper]{article}

\pdfoutput=1

\usepackage{cite}
\usepackage[cmex10]{amsmath}
\usepackage{amsfonts}
\usepackage{amssymb}
\usepackage{array}
\usepackage{url}
\usepackage{fixltx2e}
\usepackage{algorithm,algorithmicx,algpseudocode}
\usepackage{framed}
\usepackage[pdftex]{graphicx}
\usepackage{subfig}
\usepackage{epstopdf}
\usepackage{amsthm}
\usepackage{bm}
\usepackage{multirow}
\usepackage{epsfig}
\usepackage{pdflscape}
\usepackage{rotating}
\usepackage{float}
\usepackage{comment}
\usepackage[backref=page,hidelinks]{hyperref}
\usepackage[affil-it]{authblk}
\usepackage[title]{appendix}

\newtheorem*{definition}{Definition}
\newtheorem*{theorem}{Theorem}

\def\RR{\mathbb{R}}

\newcommand{\abs}[1]{\left\vert#1\right\vert}

\newcommand{\prt}[1]{\left(#1\right)}

\usepackage{color}
\definecolor{orange}{rgb}{1,0.5,0}

\newcommand{\ST}{\mathcal{SH}}

\newcommand{\ki}{i}
\newcommand{\kk}{k}

\begin{document}

\title{Scale Invariant Interest Points with Shearlets}

\date{}

\author[1]{Miguel~A.~Duval-Poo}
\author[1]{Nicoletta~Noceti}
\author[1]{Francesca~Odone}
\author[2]{Ernesto~De~Vito}
\affil[1]{Dipartimento di Informatica Bioingegneria Robotica e Ingegneria dei Sistemi (DIBRIS), Universit\`{a} degli Studi di Genova, Italy}
\affil[2]{Dipartimento di Matematica (DIMA), Universit\`{a} degli Studi di Genova, Italy}

\maketitle

\begin{abstract}
Shearlets are a relatively new directional multi-scale framework for signal analysis, which have been shown effective to enhance signal discontinuities such as edges and corners at multiple scales. In this work we address the problem of detecting and describing blob-like features in the shearlets framework. We derive a measure which is very effective for blob detection and closely related to the Laplacian of Gaussian. We demonstrate  the measure satisfies the perfect scale invariance property in the continuous case. In the discrete setting, we derive algorithms for blob detection and keypoint description. Finally, we provide qualitative justifications of our findings as well as a quantitative evaluation on benchmark data. We also report an experimental evidence that our method is very suitable to deal with compressed {and noisy} images, thanks to the sparsity property of shearlets.
\end{abstract}

\section{Introduction}

Feature detection consists in the extraction of perceptually interesting low-level features over an image, in preparation of higher level processing tasks. In the last decade a considerable amount of work has been devoted to the design of effective and efficient local feature detectors able to associate with a given interesting point also scale and orientation information. Scale-space theory has been one of the main sources of inspiration for this line of research, providing an effective framework for detecting features at multiple scales and, to some extent, to devise scale invariant image descriptors.

In this work we refer in particular to blob features, image regions which are approximately uniform. In early works the Laplacian of the Gaussian (LoG) operator has been proposed as a way of enhancing blob-like structures \cite{lindeberg1998feature}. Later, difference of Gaussians (DoG) has been introduced as an efficient approximation of the Laplacian \cite{lowe2004distinctive},  while the  Hessian determinant \cite{lindeberg1998feature} was suggested as an alternative operator with a higher sensitivity and better invariance properties. Later on, computationally efficient variants have also been devised \cite{bay2008speeded,agrawal2008censure}. Since feature detection often precedes feature matching, local features need to be associated with an appropriate descriptor. For a reliable feature matching, it is important to identify a descriptor able to deal with geometric transformations, illumination changes, and the presence of noise. Therefore over the years there has been  a lot of work in devising feature descriptors able to address different types of variations {\cite{mikolajczyk2005performance,ke2004pca,lowe2004distinctive,bay2008speeded,tola2010daisy,wang2011local}}. It should be noticed how, in this context, much effort has been devoted to reducing the computational cost --- it is worth mentioning the well known SIFT \cite{lowe2004distinctive}  and SURF \cite{bay2008speeded} feature descriptors, obtained  from scale-normalized derivatives in a scale-space representation based on image pyramids to speed up computation. More recently compact representations have been obtained by means of binary descriptors \cite{calonder2010brief,leutenegger2011brisk,alahi2012freak}.

{Unsurprisingly, image feature detection at multiple scales has also been addressed in the context of wavelet theory\cite{mallat1992singularity,mallat1992characterization,chan1992corner,chen1995wavelet,pedersini2000multi,fauqueur2006multiscale,damerval2007blob}. This framework allows for a natural derivation of the feature scale \cite{ mallat1992singularity,damerval2007blob} and for the design of perfect scale-invariant measurements \cite{fuhr2012continuous}. It is equivalent to the scale-space representation it the mother wavelet is the derivative of the Gaussian \cite{ mallat1992singularity}, but  it also allows for the choice of different mother functions  to better enhance specific features.}

{While for 1D signals, wavelets and space-scale theory are the canonical multi-scale representations, for 2D signals} there is a large class of representations with a further sensitivity to directional information, useful to deal with rotation invariance --- here it is worth mentioning directional wavelets \cite{antoine1996two}, contourlets \cite{po2006directional}, complex wavelets \cite{selesnick2005dual}, ridgelets \cite{candes1999ridgelets}, curvelets \cite{candes2004new}, and shearlets \cite{easley2008sparse}.

{In this paper we focus on  shearlets representation and we show
  how local extremas of the shearlet coefficients may enhance
  blob structures in an image providing \cite{lindeberg2015image},
 \begin{itemize}
 \item[a)] a clear  definition of these interest points;
\item[b)] a well-defined position in image space;
\item[c)] a local image structure with a rich directional information
  content;
\item[d)] a stable procedure with an high degree of repeatibillity
  against   noise and deformations;
\item[e)] the capability to detect other interesting point, like edges and cornes,
with a different choice of the generating function  \cite{duval2015edges};
\item[f)] an automatic scale selection with a scale invariant descriptor.
 \end{itemize}
}

Indeed, shearlets  enjoy different interesting properties which are meaninful to feature detection and description:
\begin{itemize}
\item As in the space-scale approach, the filters give rise to a coarse-to-fine multi-scale representation, but for shearlets two consecutive scales are related by an {anisotropic dilation with ratio $1/\sqrt{2}$} and this anisotropy is the key property to have (optimal) sparse representation for natural images \cite{labate2005sparse}. Among the multi-scale representations, only shearlets and curvelets ensure this kind of optimality, so that shearlets are appealing in applications dealing with signal compression.
\item The shearlet coefficients directly encode directional information, unlike scale-space representations and traditional wavelets where one could derive directional information only as a post-processing, for example by computing the ratio between the partial derivatives.
\item The rotational invariance of the representation is given by a shearing which preserves the rectangular lattice of the digital image, so that a faithful digital implementation is easy to obtain.

\item In contrast to the scale-space approaches, with shearlets we have a large choice of admissible templates allowing to tune the shearlet transform to specific applications, e.g, the Gaussian derivative to locate edges or corners as in \cite{yi2009shearlet, duval2015edges}, or the Mexican hat to analyze blob structures or ridge points.
\item Shearlets also appear to have a potential in providing meaningful descriptions, although this capability has not been largely explored so far (see, for instance,   \cite{schwartz2011novel,he2013rotation}).
\end{itemize}

In this paper, first,  we provide an analysis of  perfect scale invariance properties in the continuous case, similar to the study carried out by  Lindberg for the scale-space {\cite{lindeberg2013scale}}. Then, we derive a discretized formulation of the problem, obtaining a discrete measure which will be the main building block of our algorithms. This measure, obtained by summing up coefficients over all the shearing, is naturally isotropic, but we can easily recover the directional information by looking at the single coefficient.

Next, we propose an algorithm for detecting and describing blob-like features. The main peculiarity of our approach is in the fact it fully exploits the expressive power of the shearlet transform. Indeed, each detected feature is associated with a scale, an orientation, and a position, directly related with the dilation, the shearing and the translation provided by the underlying transformation. In the description phase we also use shearlets coefficients, orienting the contributions with respect to the estimated feature orientation. We underline how all the steps of our procedure are based on the same image transformation. In this sense the procedure is elegant and clean and has a potential in computational efficiency.

We present a comparative analysis on benchmark data where we show that the proposed method compares favorably with the state of the art of scale-space features. We also present a further experiment on a larger set of images, where we underline the appropriateness of the method to address image matching at different compression {and noise} levels. In this specific aspect resides one of the main contribution of our work from the application standpoint: the sparsity properties of the shearlet transform are very appropriate to deal with noise and compression artifacts.

The paper is organized as follows: In Section \ref{sec:shear_CST} we review the shearlet transform. Section \ref{sec:shear_scale_selection} provide an analysis of scale selection in multi-scale image representations and the theoretical justifications of scale invariance for feature detection by shearlets. In Section \ref{sec:sbd} we propose the shearlet based blob detection algorithm, while the descriptor is introduced in Section \ref{sec:shear_desc}. Section \ref{sec:blob_exp_results} reports a experimental analysis of the proposed blob detector and descriptor following the Oxford evaluation procedure. In Section \ref{sec:bdd_compress_image_exp} we evaluate the proposed methods for image matching at different compression {and noise} levels. Section \ref{sec:bdd-shear_discussion} is left to a final discussion.

\section{A Review of the Shearlet Transform}
\label{sec:shear_CST}

A shearlet is generated by the dilation, shearing and translation of a function $\psi \in L^2(\RR^2)$, called the {\em mother shearlet}, in the following way
\begin{equation}
\label{eq:psi_cont_def}
\psi_{a,s,t}(x)= a^{-3/4}\psi(A^{-1}_a S^{-1}_s(x-t))
\end{equation}
\noindent where $t\in\RR^2$ is a translation, $A_a$ is a \emph{dilation} matrix and $S_s$ a \emph{shearing} matrix defined respectively by
\begin{equation*}
A_a=\begin{pmatrix}a& 0 \\ 0&\sqrt{a} \end{pmatrix} \qquad S_s = \begin{pmatrix}1&   s \\ 0&1 \end{pmatrix},
\end{equation*}
with $a\in \RR^+$ and $s\in \RR$. The anisotropic dilation $A_a$
controls the scale of the shearlets, by applying a different dilation
factor along the two axes. The shearing matrix $S_s$, not expansive,
determines the orientation of the shearlets. The normalization factor
$a^{-3/4}$ ensures that $\|\psi_{a,s,t}\|=\|\psi\|$, where $\|\psi\|$
is the norm in $L^2(\RR^2)$. The \emph{shearlet transform} ${\ST} (f)$
of a signal $f\in L^2(\RR^2)$ is defined by
\begin{equation}
\label{eq:cont_shear_trans}
    {\ST}(f)(a,s,t) = \langle f,\psi_{a,s,t}\rangle
\end{equation}
\noindent where
$\langle f,\psi_{a,s,t}\rangle$ is the scalar product in $L^2(\RR^2)$.

In the classical setting the \emph{mother shearlet} $\psi$ is assumed to factorize in the Fourier domain as
\begin{equation}
\label{eq:psi_factorization}
\hat{\psi}(\omega_1,\omega_2)=\hat{\psi}_1(\omega_1)\hat{\psi}_2(\frac{\omega_2}{\omega_1})
\end{equation}
where $\hat{\psi}$ is the  Fourier transform of $\psi$, $\psi_1$ is a
one dimensional wavelet and $\hat{\psi_2}$ is any non-zero
square-integrable function.

\begin{figure}[!t]
   \centering
     \includegraphics[width=0.5\textwidth]{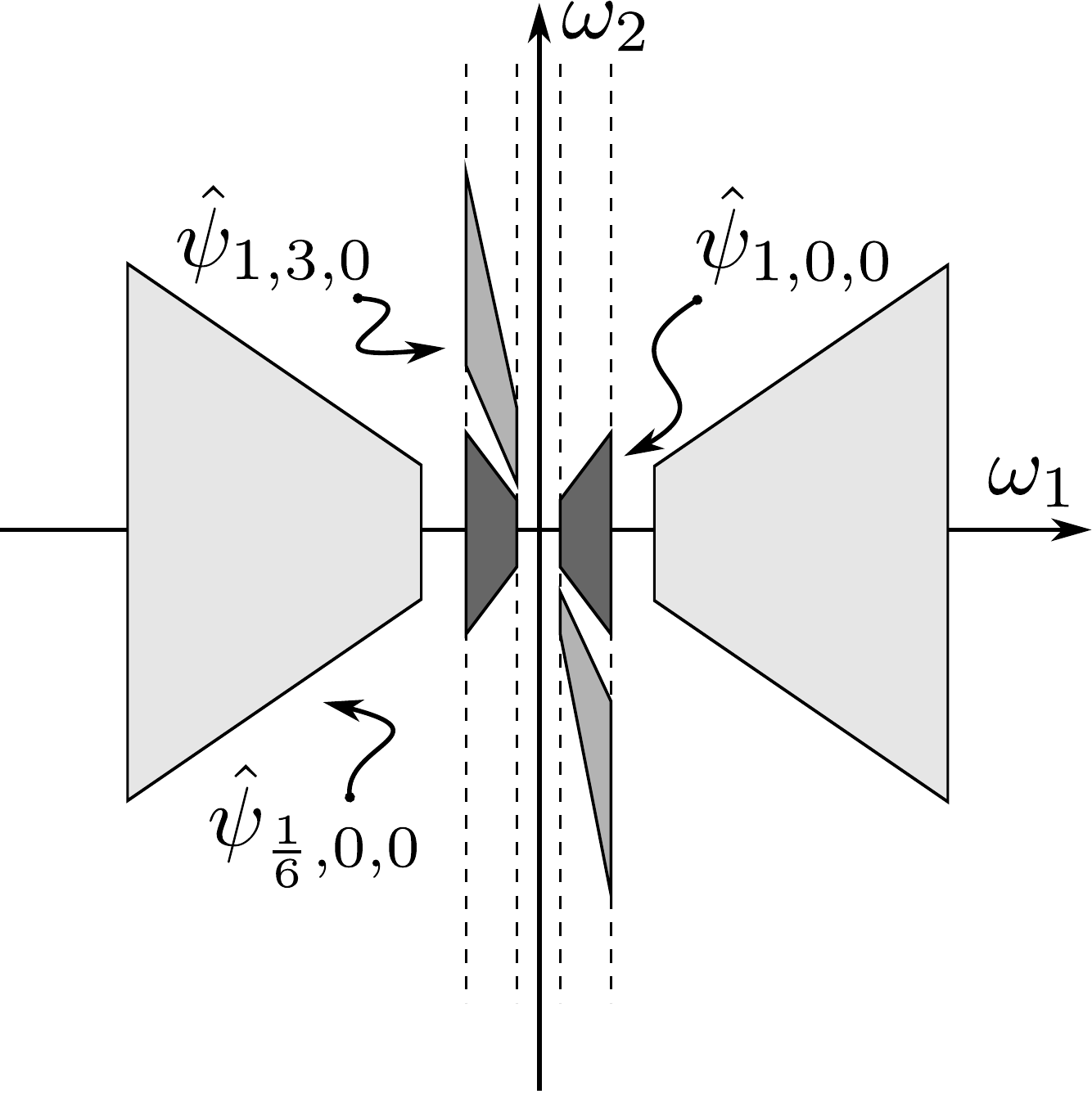}
   \caption{Support of the shearlets $\hat\psi_{a,s,t}$ (in the frequency domain) for different values of $a$ and $s$.}
   \label{fig:shear_support}
 \end{figure}

With the choice of Eq. \eqref{eq:psi_factorization} the shearlet definition in the frequency domain {(see Fig. \ref{fig:shear_support})} becomes
\begin{equation*}
\hat{\psi}_{a,s,t} (\omega_1,\omega_2)=
a^{3/4}\hat{\psi}_1(a\omega_1)\hat{\psi}_2\prt{\frac{\omega_2+s\omega_1}{\sqrt{a}\,\omega_1}}e^{-2\pi i (\omega_1,\omega_2) \cdot t }.
\end{equation*}

 \graphicspath{{./images/ch-shearlets/}}

\noindent As a consequence of  the Plancherel formula, Eq. (\ref{eq:cont_shear_trans}) can be rewritten as
\begin{multline}\label{eq:cst}
        {\ST}(f)(a,s,t_1,t_2) = a^{3/4}
        \int\limits_{\widehat{R}^2}\hat{f}(\omega_1,\omega_2)\hat{\psi}_1(a\omega_1)\\
        \times\hat{\psi}_2\prt{\frac{\omega_2+s\omega_1}{\sqrt{a}\,\omega_1}}
        e^{2 \pi i \xi \left( t_1 \omega_1+ t_2\omega_2\right)} d\omega_1d\omega_2.
\end{multline}

A different approach is proposed in \cite{kittipoom2012construction}
where the mother shearlet is of the form $\psi_1(x_1)\psi_2(x_2)$,
where $\psi_1$ is a one dimensional wavelet and $\psi_2$ is a scaling
function. This property allows to have compact support shearlets in
the space domain.  However, as noted in \cite{kutyniok2014shearlab},
{\em `` The property [of~\eqref{eq:psi_factorization}] does indeed not only
improve the frame bounds of the associated system, but also improves
the directional selectivity significantly''}.  To overcome this
problem, in \cite{kutyniok2014shearlab} the mother shearlet is
multiplied by a suitable 2D fan filter in order to approximate
the property~\eqref{eq:psi_factorization}. For sake of simplicity, in
this short review we consider only classical shearlets --- the ones we have adopted in our work.

\subsection{Cone-adapted Shearlets}
\label{sec:shear_cone_adapted}

A major limitation of the shearlets defined in the previous section is the directional bias of shearlet elements associated with large shearing parameters.
To deal with this problem  \cite{labate2005sparse}  introduces
the notion of  \emph{cone-adapted shearlets} whose construction is
based on a partition of the Fourier into two cones and a square centered around the origin.
The two conic regions are defined as
\begin{align*}
\mathcal{C}_\mathrm{h} &= \{ (\omega_1,\omega_2)\in\RR^2 : |\omega_2 / \omega_1| \leq 1, |\omega_1| > 1 \} \\
\mathcal{C}_\mathrm{v} &= \{ (\omega_1,\omega_2)\in\RR^2 : |\omega_1 / \omega_2| \leq 1, |\omega_2| > 1 \}.
\end{align*}
\noindent  A shearlet $\psi$ suitable for the horizontal cone is
\begin{equation*}
\hat\psi^\mathrm{h}(\omega_1,\omega_2)=\hat{\psi}_1(\omega_1)\hat{\psi}_2\prt{\frac{\omega_2}{\omega_1}}\chi_{\mathcal{C}_\mathrm{h}}(\omega_1,\omega_2).
\end{equation*}
where  $\chi_{\mathcal{C}_\mathrm{h}}(\omega)$  is equal to 1 for
$\omega \in \mathcal{C}_\mathrm{h}$ and 0  outside. Likewise the
shearlet for the vertical cone is defined by interchanging roles of
$\omega_1$ and $\omega_2$.

\noindent The square region is the low-frequency part
\begin{equation*}
\{ (\omega_1,\omega_2)\in\RR^2 : |\omega_1|,|\omega_2| \leq 1 \}.
\end{equation*}
\noindent Since the interest points of an image are associated with high
frequencies, for space reason we do not consider the low-frequency
contribution, see \cite{labate2005sparse}  for further details.

\subsection{Digital Shearlets}
\label{sec:shear_digital}

Digital shearlet systems are defined by sampling continuous shearlet
systems on a discrete subset of the space of parameters
$\RR_+\times\RR^3$ and by sampling the signal on a grid.
In the literature there are many different discretization schemes, see
\cite{kutyniok2012shearlab,kutyniok2014shearlab} and reference
therein. In this work we adopt the Fast Finite Shearlet Transform
(FFST) \cite{hauser2014fast} which performs the entire shearlet
construction in the Fourier domain. It is possible to choose as $\Psi_1$
wavelets whose analytic form is given in the Fourier domain,  whereas
in \cite{yi2009shearlet} and \cite{kutyniok2014shearlab} are
restricted to wavelets associated with multiresolution analysis.
In this scheme, the signal is discretized on a square grid of size $N$, which is independent of the dilation and shearing parameter, whereas the  scaling, shear and translation parameters are discretized as
\begin{align*}
a_j &= 2^{-j}, \quad j=0,\dots,j_0 -1, \\
s_{j,\ki} &= \ki2^{-j/2},\quad -\lfloor 2^{j/2} \rfloor\leq \ki \leq \lfloor 2^{j/2}\rfloor, \\
t_m &=\prt{\frac{m_1}{N},\frac{m_2}{N}},\quad m\in \mathcal{I}
\end{align*}
\noindent where $j_0$ is the number of considered scales and
$\mathcal{I}=\{(m_1,m_2):m_1, m_2=0,\dots,N-1\}$. With respect to the
original implementation we use a dyadic scale $2^{-j}$ instead of
$4^{-j}$ to reduce the difference among two consecutive scales,
which is consistent with the discretization lattice in
\cite{kutyniok2014shearlab}.

 With these notations the shearlet system becomes
\begin{equation*}
\psi^{\mathrm{x}}_{{j,\ki,m}}(x) =
\psi^{\mathrm{x}}_{a_j,s_{j,\ki},t_m}(x)
\end{equation*}
\noindent where $\mathrm{x}=\mathrm{h}$ or $\mathrm{x}=\mathrm{v}$
and  the discrete shearlet transform of an digital image $\mathcal{I}$ is
\begin{equation*}
{\ST}(\mathcal{I})(j,\mathrm{x},\ki,m)=\langle \mathcal{I},\psi^\mathrm{x}_{j,\ki,m} \rangle
\end{equation*}
\noindent where $j=0,\dots,j_0-1$, $\mathrm{x}=\mathrm{h},\mathrm{v}$,
 $|\ki|\leq \lfloor 2^{j/2}\rfloor$, $m\in \mathcal{I}$. Based on the
 Plancherel formula $\langle f,g \rangle=\frac{1}{N^2}\langle \hat f,
 \hat g \rangle$, the discrete shearlet transform can be
 computed by applying the 2D Fast Fourier Transform (\texttt{fft}) and
 its inverse (\texttt{ifft}). For example, for the horizontal cone
 ${\ST}(\mathcal{I})(j,\mathrm{h},\ki,m) $ is given by
\begin{equation}
\label{eq:disc_shear_trans}
2^{-\frac{3j}{4}}\texttt{ifft}(\hat\psi_1(2^{-j}\omega_1)\hat{\psi_2}(2^{j/2}\frac{\omega_2}{\omega_1}+\ki)\texttt{fft}(\mathcal{I}))(m).
\end{equation}

\subsection{Shearing and Orientation}
\label{sec:shear_indexing}

\graphicspath{{./images/ch-shearlets/}}
\begin{figure}[!t]
  \centering
  {\includegraphics[width=0.5\textwidth]{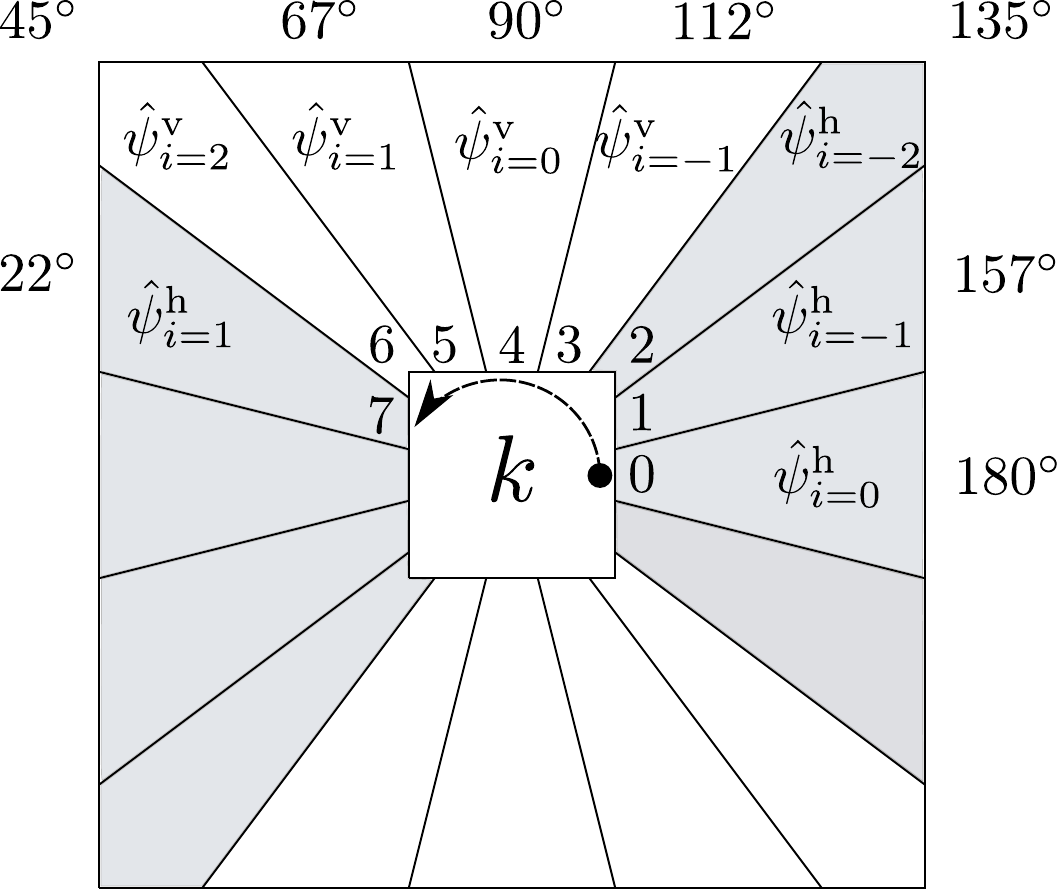}}
  \caption{A visualization, in the Fourier domain {for scale $j=2$}, of the shearing indices $i$ on the horizontal and vertical cone (gray and white areas respectively), the complete ordering of index $k$, and the associated angles $\theta_k$.}
  \label{fig:shearing_idx}
\end{figure}
As we can notice, all the shearing associated with a scale $j$ are
distributed along the two vertical and horizontal cones ${\mathrm x}$. For the sake of clarity,  in
order to provide a simple access to a specific shearing at a scale $j$,
we introduce an index $k$ (similar to \cite{hauser2014fast}) which
replaces the shearing parameter $\ki$ and the cone parameters
${\mathrm x}=\{\mathrm{h},\mathrm{v}\}$ and simplifies the notation.
For each scale $j$, the index $k$  iterates counter-clockwise the shearlets in
the Fourier domain. It starts in the horizontal cone  for each $\ki=-1, \ldots, -\lfloor 2^{j/2}\rfloor$
starting from  $\ki=0$. Then, it
continues iterating the vertical cone for $\ki = -\lfloor 2^{j/2}\rfloor
+ 1, \ldots, \lfloor 2^{j/2}\rfloor$. Once the vertical cone is completed,
it starts on the remaining of the horizontal cone, $\ki =
\lfloor 2^{j/2}\rfloor - 1, \ldots, 1$.
Figure \ref{fig:shearing_idx} illustrates  the relationship between the original index $\ki$ and $k$.
Hence, from now on, we will consider the following formulation of the discrete shearlet transform
\[
{\ST}(\mathcal{I})(j,\kk,m)= {\ST}(\mathcal{I})(j,\mathrm{x},\ki,m)
\quad \kk = 0,\ldots, 4\lfloor 2^{j/2}\rfloor {-1},\]
where $4\lfloor 2^{j/2} \rfloor$ is the total number of shearlets for a scale $j$.

Now, we may associate an orientation $\theta_k$ to each index $k$:
\begin{equation}
\label{eq:theta_k}
\theta_k = \pi\left(1-\frac{k}{4\lfloor 2^{j/2} \rfloor}\right).
\end{equation}

\section{Scale Selection with Shearlets}
\label{sec:shear_scale_selection}

Multi-scale frameworks, like scale-space, wavelets and shearlets, represent image structures at multiple scales and are thus appropriate for detecting structures or features with different spatial extent. They may also be effective in estimating an appropriate scale to a given detected feature, useful for further tasks such as matching or recognition.
According to Lindeberg \cite{lindeberg2014scale}, the formal definition of \emph{scale selection} refers to the estimation of characteristic scales in image data and the automatical selection locally appropriate scales in a scale-space representation.

A particularly useful methodology for computing estimates of characteristic scales is by detecting local extrema over scales of differential expressions in terms of $\gamma$-normalized derivatives \cite{lindeberg1998feature}. Following this approach, it can be shown that different types of scale invariant feature detectors can be used for different types of visual features, like blobs, corners, etc. Furthermore, the scale levels obtained from the scale selection can be used for computing image descriptors.

Detected feature points are considered \emph{scale invariant} if the points are preserved under scaling transformations and the selected scale levels are transformed in accordance with the amount of scaling \cite{lindeberg1998feature}.
In addition, \emph{perfect scale invariance} is considered when the extrema over scales are equal.

In this section, we start from the continuous setting and first discuss scale invariance properties
reviewing Lindeberg approach in the framework of space-scale
theory \cite{lindeberg1998feature}. Then, we show how shearlet coefficients can also detect the correct
scale while providing directional information. In the second part of the section we discuss how we can obtain a measure of scale in the discrete setting.

\subsection{Scale Invariance in the Continuous Setting}

When a signal is subject to a scale-space smoothing, the spatial
derivatives on the smoothed data are expected to decrease. This is a
well-known property of the scale-space representation, according to
which the amplitude of its spatial derivatives decreases with
scale. To obtain a multi-scale signal representation whose amplitude
is independent on the scale Lindeberg proposed a $\gamma$-normalized derivative
operator \cite{lindeberg1998feature}. In this section, we will show
that shearlets share a similar behaviour, {but they
  directly encode the directional information}. We observe how a similar analysis could be carried out considering directional
wavelets \cite{antoine1996two}, which are not included in our
study but will be taken into account in future works.

\subsubsection{Scale-space} {In this section we briefly recall the
  main properties of  scale-space theory for two dimensional
  signals developed by Lindeberg \cite{lindeberg1998feature}.

In the scale-space theory, the filters are given by the family of 2D-Gaussian kernels
\[
g_a(x)=\frac{1}{2\pi a^2}e^{-\frac{x^2}{2a^2}} \qquad x\in\mathbb R^2
\]
parametrised by the  scale $a\in\mathbb R^+$.  Each signal $f$ is
mapped to its scale-space transform by convolution with $g_a$
\begin{align*}
L[f](a,z)&=g_a\ast f(a z)=\int_{\RR^2} g_a(y)f(a z-y)dy\nonumber
\end{align*}
where $z=x/a$ is the ``scale invariant'' space variable. In the original paper
\cite{lindeberg1998feature} the scale  is $t=a^2$ and a more
general rescaling is considered ( as for example $z=x/t^{\gamma/2}$). This
representation has two main properties, which are at the root of the
theory and are usually referred as perfect scale invariance.  First, the representation $L[f]$ is invariant under
dilations, {\em i.e.} if $f_\alpha(x)=f(x/\alpha)$ with
$\alpha\in\RR^+$, then
\[
L[f_\alpha](a,z) = L[f](\alpha^{-1}a,z).
\]
Furthermore,   for 2D sinusoidal signals
\begin{equation}
\label{eq:2d_sin}
f(x_1, x_2) = \cos(\alpha x_1) \cos(\beta x_2),
\end{equation}
the corresponding transform is able to detect the scale by applying a
suitable differential operator
\[
\mathcal D= P(\frac{\partial}{\partial z_1},\frac{\partial}{\partial z_2})
\]
where $P$ is  a given polynomial in two variables. Indeed, define the
quantity
\[  L_{\mathcal D,\max}(a)=\max_{z\in\RR^2} |\mathcal DL[f](z,a)|, \]
since
\[
L[f](a,z)= e^{-\frac{a^2(\alpha^2+\beta^2)}{2}} f(az),
\]
then
\begin{equation}
  \label{eq:max_2d_scalespace}
 L_{\mathcal D,\max}(a)= P(a\alpha,a\beta)  e^{-\frac{a^2}{2}(\alpha^2+\beta^2)}.
\end{equation}
For example, if $\mathcal D=\frac{\partial^2}{\partial z_1^2}+ \frac{\partial^2}{\partial
  z_2^2}$ is the $z$-Laplacian, then
\[
L_{\mathcal D,\max}(a)= a^2(\alpha^2+\beta^2) e^{-\frac{a^2}{2}(\alpha^2+\beta^2)},
\]
which takes its maximum  at ${a^*=1/\sqrt{\alpha^2+\beta^2}}$ with value
independent on the scale.  Hence, the extrema of the scale-space
representation across the scale $a$ allows to detect the scale
$1/\sqrt{\alpha^2+\beta^2}$  of the signal. However, to extract the ratio
$\beta/\alpha$ associated with the directional information there is
the need to compute other quantities as the determinant of the
Hessian \cite{lindeberg2015image}. In the next section we show that
shearlets have essentially the same behaviour, but the transform
directly detect both the parameters $\alpha$ and $\beta$.
}

\subsubsection{Shearlets}
Since the dilation matrix defining the shearlets  is not isotropic,
we can not expect that the shearlet transform itself is invariant
under (isotropic) scale changes. However, we will show how a related
quantity has the perfect scale invariance property, as demonstrated by
the following result, whose proof can be found in the appendix.
\begin{theorem}\label{th:one}
The rotationally invariant shearlet transform
\begin{equation}
\label{eq:cont_blobness}
B[f](a,z) =a^{-5/4} \int_{\mathbb{R}} \mathcal{SH}(f)(a,s,az)ds,
\end{equation}
with $a\in\mathbb R_+$ and $z\in\mathbb R^2$, is scale invariant, {\em
  i.e.} for all $f\in L^2(\RR^2)$
\begin{equation*}
B[f_\alpha](a,z) = B[f](\alpha^{-1}a,z ).
\end{equation*}
Furthermore, if $f$ is the  sinusoidal signal given
by~\eqref{eq:2d_sin}, then
\begin{equation*}
B[f](a,z) = \overline{\psi_2(0)} \overline{\hat{\psi_1}(\frac{a\alpha}{2\pi})}f(
az),
\end{equation*}
provided that $\Psi_1$ is even.
\end{theorem}

\begin{figure}[!t]
  \centering
  \subfloat[Scale-space]{
  \includegraphics[width=0.4\textwidth]{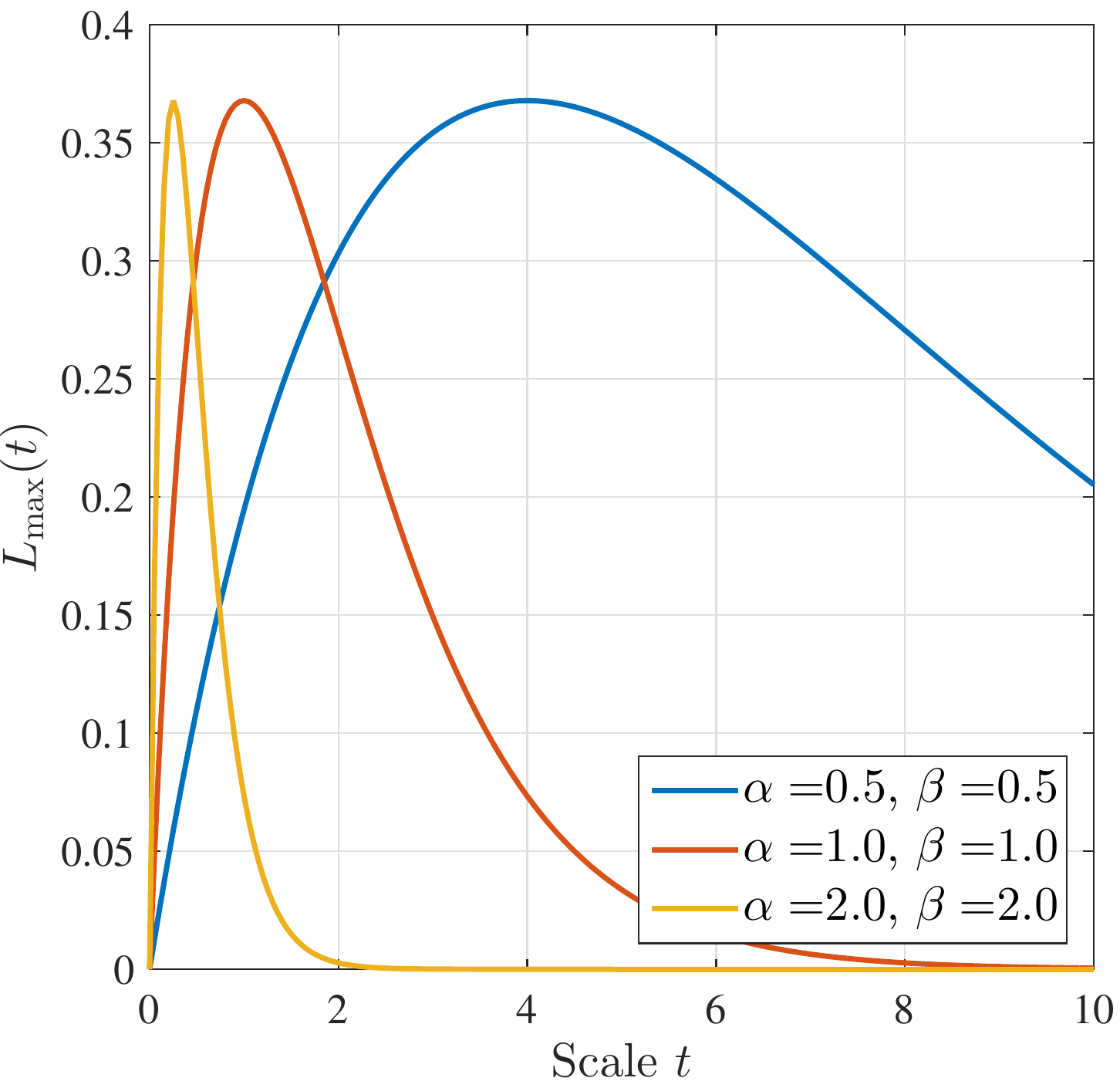}\
  \includegraphics[width=0.4\textwidth]{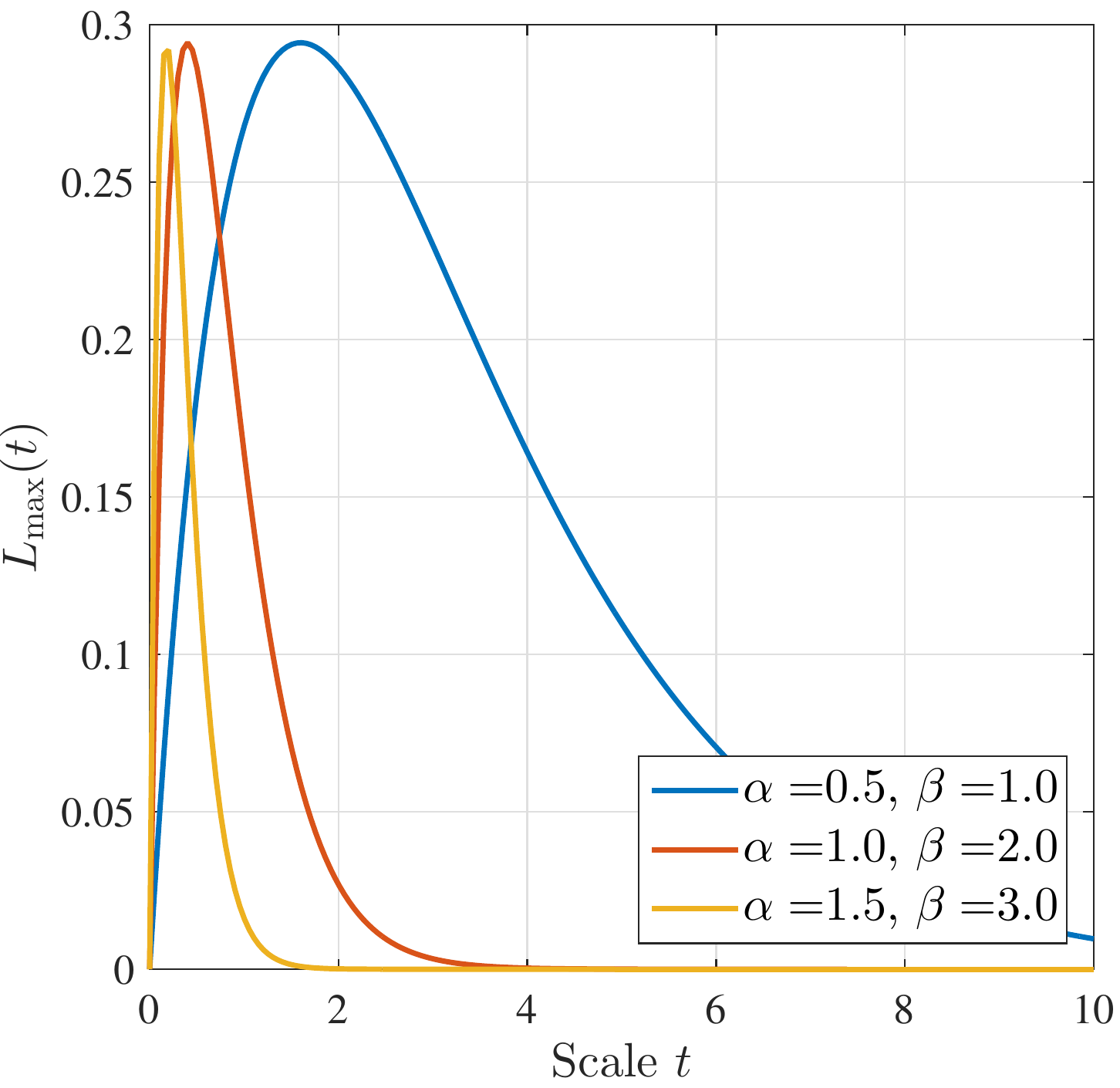}}\\
  \subfloat[Shearlets]{
  \includegraphics[width=0.4\textwidth]{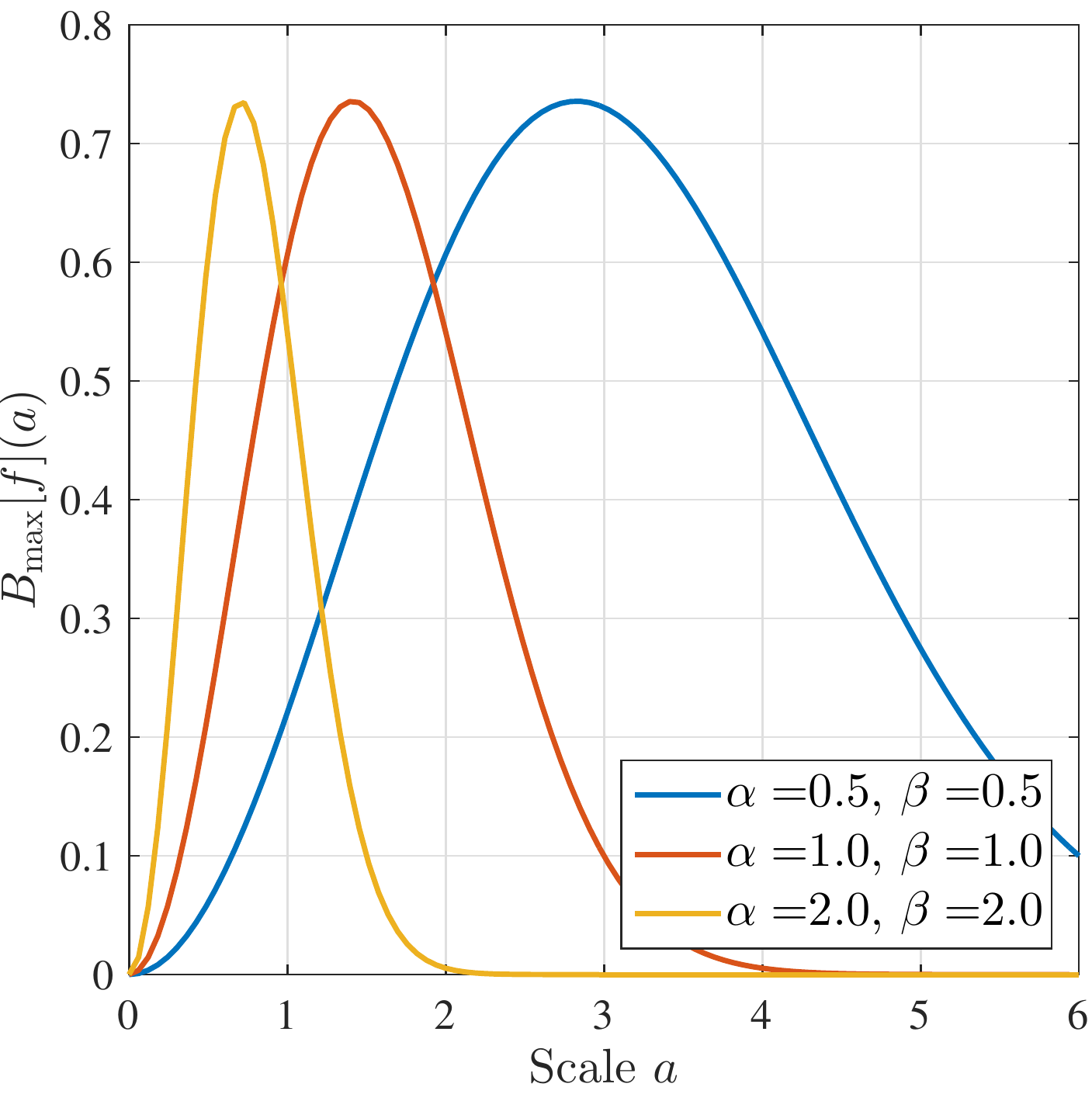}\
  \includegraphics[width=0.4\textwidth]{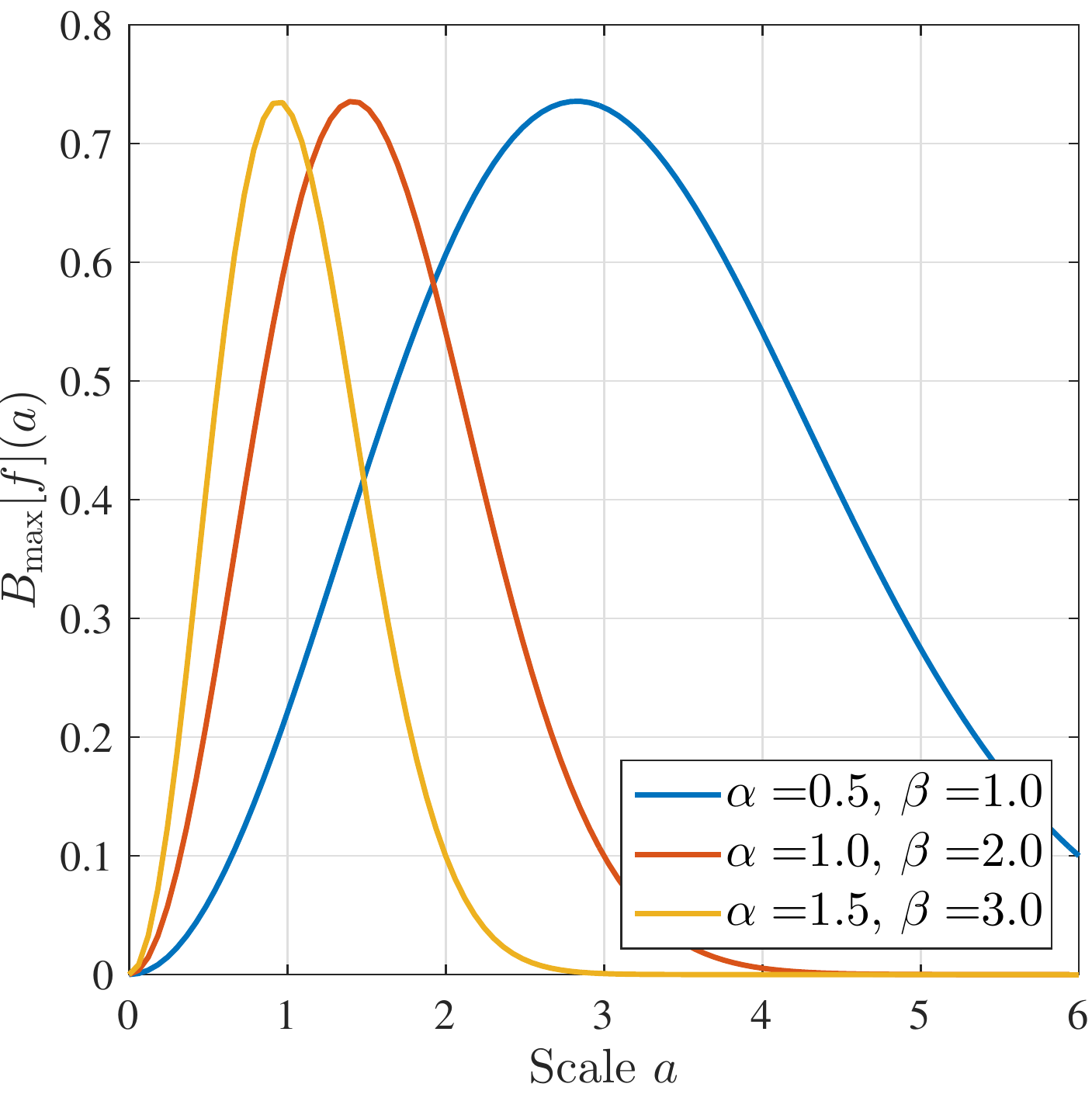}}
  \caption{
  The plot of (a) $L_{\max}(t)$  Eq. \eqref{eq:max_2d_scalespace} and (b)
  {$B_{\max}[f](a)$} Eq. \eqref{eq:max_2d_shearlets} as function of scale for 2D
    sinusoidal signals at different frequencies $\alpha$ and $\beta$.
  }
  \label{fig:2d_sin_analysis}
\end{figure}

As we did for scale-space, if $f$ is the  2D sinusoidal
signal as in~\eqref{eq:2d_sin} a simple calculation shows that the maximum of the modulus over $z$ is
\begin{equation}
\label{eq:binvf}
B_{\max}[f](a) = |\psi_2(0)|\,|\hat{\psi_1}(\frac{a\alpha}{2\pi})|.
\end{equation}
By choosing $\hat\psi_1$ as the $1D$-\emph{Mexican hat} wavelet
\begin{equation}
\label{eq:mex_hat_wavelet}
\hat{\psi_1}(\omega) = \omega^2e^{-2\pi^2\omega^2},
\end{equation}
\noindent we can rewrite Eq. \eqref{eq:binvf} as
\begin{equation}
\label{eq:max_2d_shearlets}
B_{\max}[f](a) = \frac{|\psi_2(0)|}{4\pi^2} (a\alpha)^2 e^{-\frac{(a\alpha)^2}{2}},
\end{equation}
which shares the same behaviour of $L_{\max}[f]$, but the {
    maximum of $B_{\max}[f]$ is now
at $a^*=1/\alpha$.  Hence, for  shearlets the selected scale  $a^*$ only depends on the
frequency $\alpha$, as shown in Fig.~\ref{fig:2d_sin_analysis}. }
However, if we consider the shearlet coefficient, a computation as above shows that
\[
\max_{t\in\RR^2} |\mathcal{SH}(f)(a,s,t)| = a^{3/4}
|\hat{\psi_1}(\frac{a\alpha}{2\pi})| |\hat{\psi}_2(\frac{s+\beta\alpha^{-1}}{\sqrt{a}})|,
\]
provided that both $\psi_1$ and $\psi_2$ are even.
Since $\hat{\psi}_2$ is a bump function, fixed the scale $a$,
the shearlet coefficients have a maximun around an interval
centered at $s=-\beta/\alpha$. If $\hat{\psi}_2$ is a Gaussian bump
and $\Psi_1$ is as in~\eqref{eq:mex_hat_wavelet}
\[
a^{-3/4}\max_{t\in\RR^2} |\mathcal{SH}(f)(a,s,t)|= C (a\alpha)^2
e^{-\frac{(a\alpha)^2}{2}} e^{-\frac{( s+\beta\alpha^{-1})^2}{2a}}.
\]

\graphicspath{{./images/ch-blobs/scale_selection/2d_sin_analysis/}}

Fig. \ref{fig:2d_sin_analysis} presents the actual plots of Eq. \eqref{eq:max_2d_scalespace} for the scale-space  and Eq. \eqref{eq:max_2d_shearlets} for shearlets  at different combinations of the frequencies parameters $\alpha$ and $\beta$. In the  plots on the left, the frequencies $\alpha$ and $\beta$ are equally increased by a factor of 2 (isotropic structures), while on the right, $\alpha$ is increased by a step equal to 0.5 and $\beta=2\alpha$ (anisotropic structures).
As we can observe, the plots associated with both approaches are capable to produce perfect scale
invariance for the two types of frequency combinations in the
sinusoidal function $f$.

\graphicspath{{./images/ch-blobs/scale_selection/2d_gauss/}}
\begin{figure*}[!ht]
  \centering
  \subfloat[2D Gaussian functions with different $\sigma_x,\sigma_y$]{
  \includegraphics[width=0.35\textwidth]{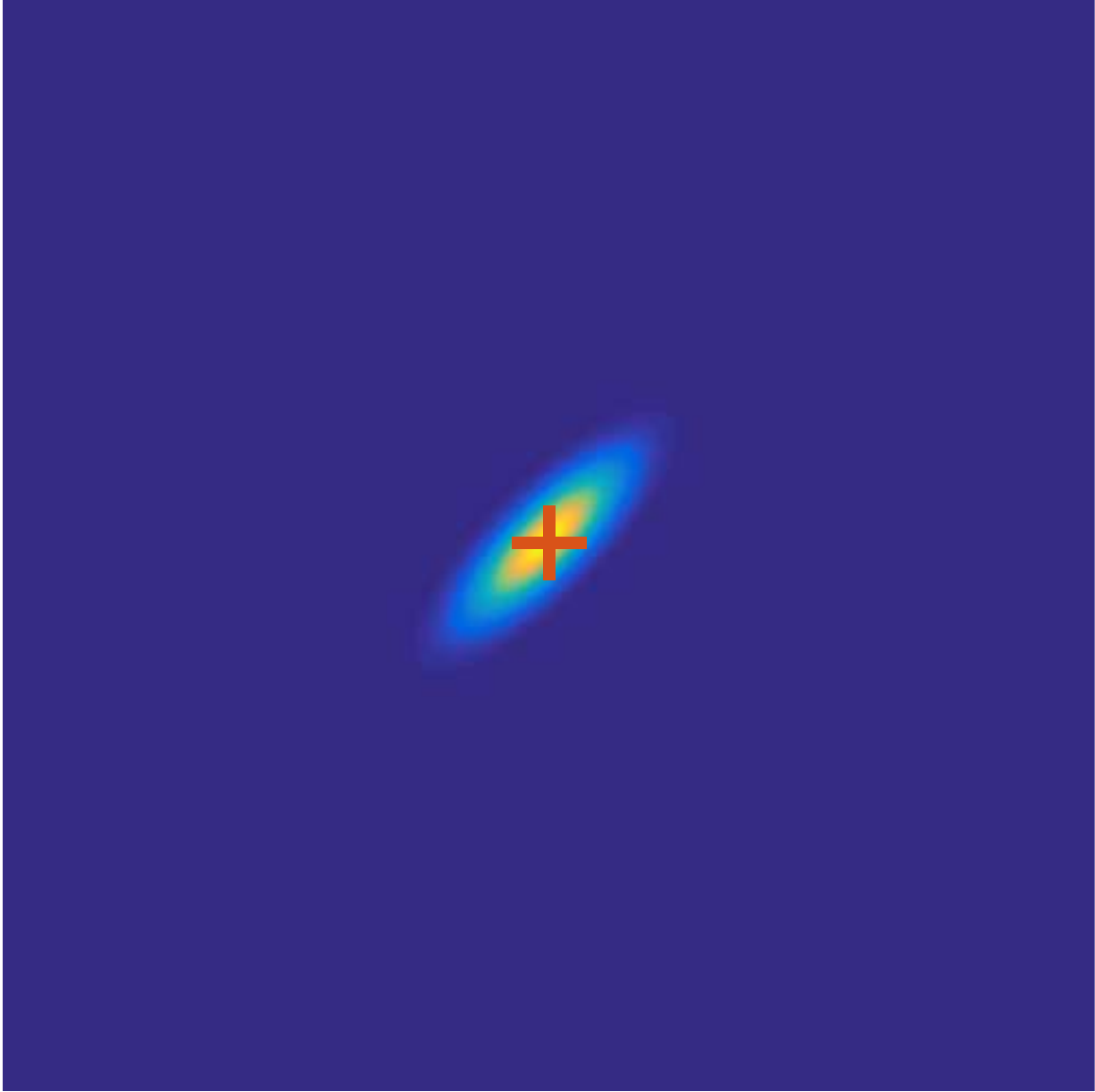}\
  \includegraphics[width=0.35\textwidth]{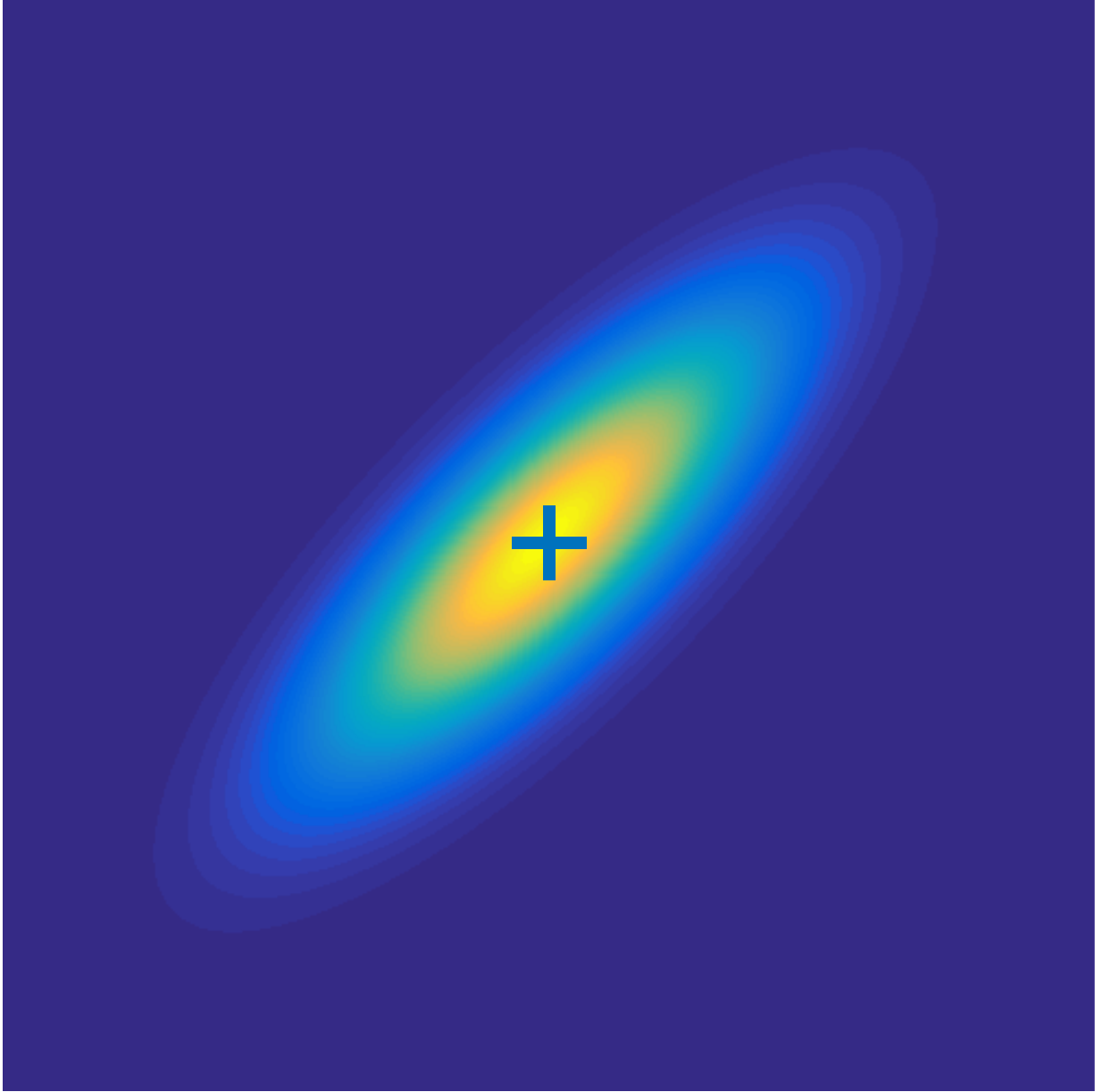}}\\
  \subfloat[Laplacian]{\includegraphics[width=0.37\textwidth]{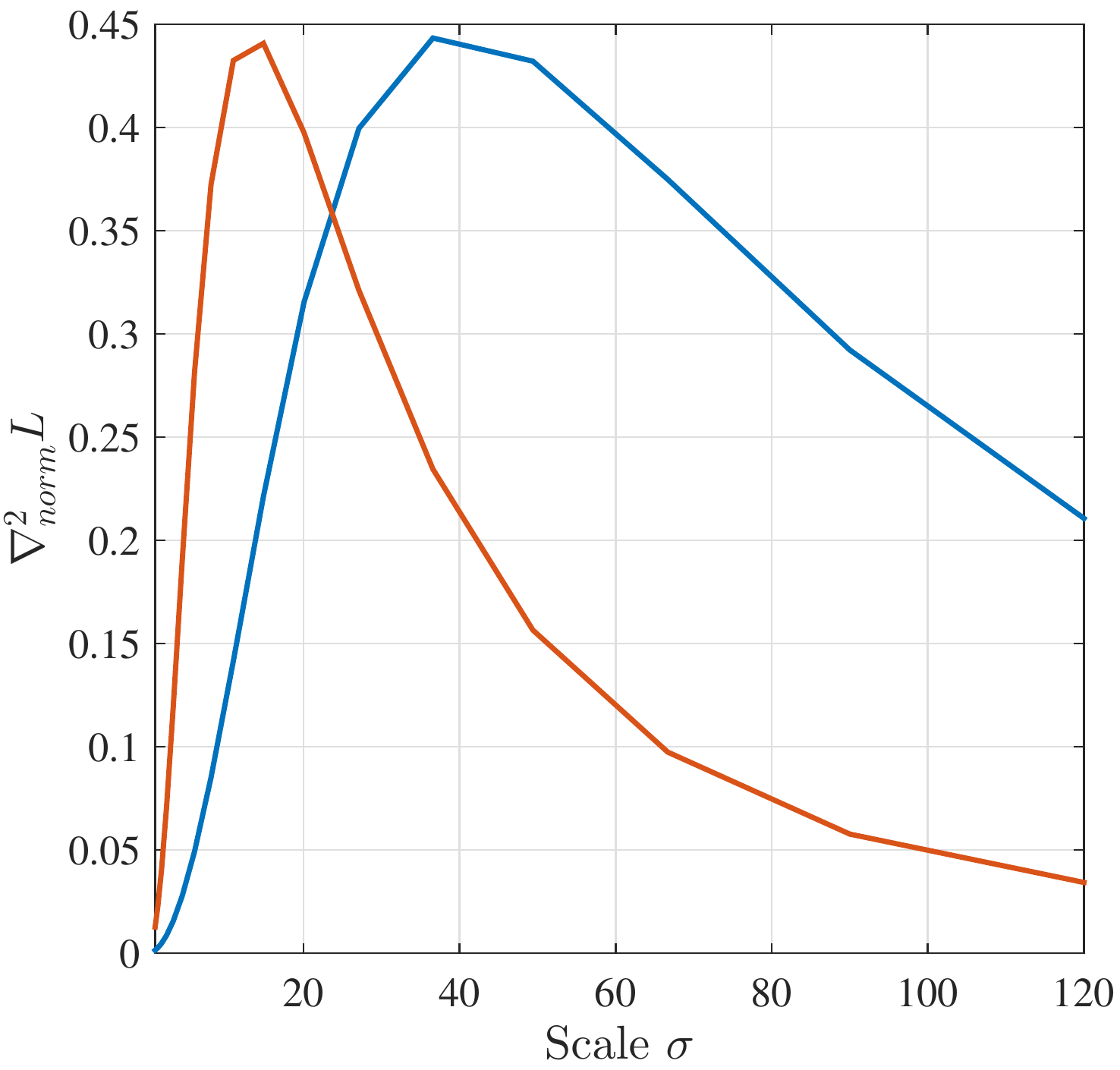}}
  \subfloat[Shearlets]{\includegraphics[width=0.37\textwidth]{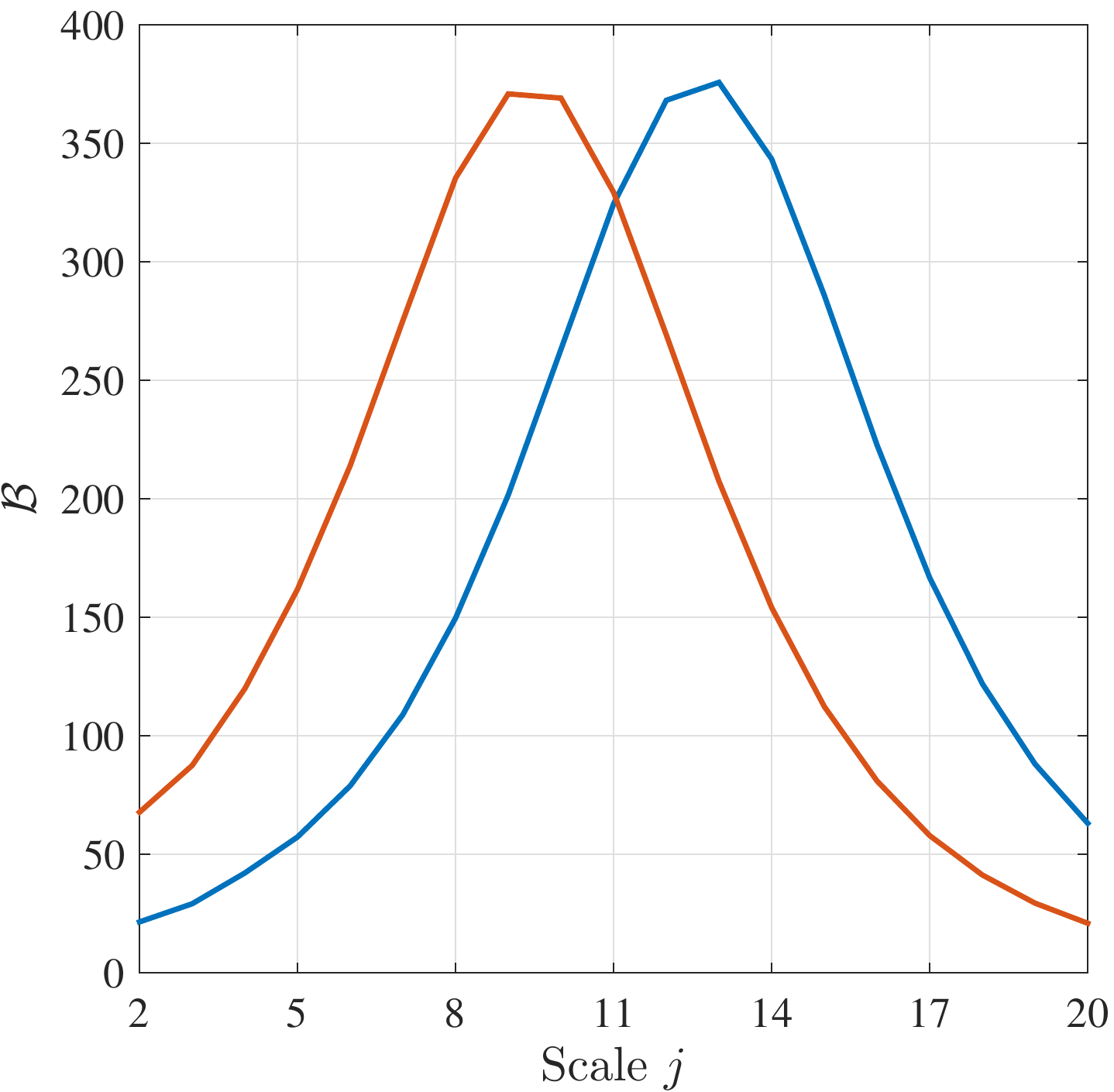}}
  \caption{Scale decomposition of the synthetic images in (a) at their center point (color-coded) by using the (b) Laplacian of Gaussian,  and (d) shearlet $\mathcal{B}$ measure.}
  \label{fig:2d_gauss}
\end{figure*}

{Finally, we stress that the choice of $\psi_1$ and $\psi_2$
  influences the type of local features that are enhanced by the
  shearlet transform. Thus, in order to detect blob features, as
  suggested by Equation~\eqref{eq:mex_hat_wavelet} we selected $\psi_1$ as the Mexican hat wavelet   and $\psi_2$ as a smooth function with compact support whose analytic form is given in \cite{hauser2014fast}.}

\subsection{Scale Invariance in the Discrete Setting}

In the previous section, we defined a scale invariant shearlet transform in the continuous setting. Now, let us formally define the discrete counterpart of Eq. \eqref{eq:cont_blobness}, which we call the $\mathcal{B}$ measure.

\begin{definition}
The $\mathcal{B}$ measure is the scale-normalized sum of the discrete shearlet transform coefficients across  the shearing parameter, 
\begin{equation}
\label{eq:blobness}
\mathcal{B}(m,j) = \frac{2^{\frac{5j}{4}}}{C_j}\sum_k \mathcal{SH}(\mathcal{I})(j,k,m),
\end{equation}
where $j, k, m$ are the discretized scaling, shearing and translation
parameters.
\end{definition}
{Comparing with Eq.~\eqref{eq:cont_blobness}, the normalization factor $C_j$ takes into account that for each
  scale $j$ there is a different number of orientations. }
We now briefly discuss the concept of perfect scale invariance on a discrete synthetic signal.
Fig. \ref{fig:2d_gauss} (a) shows two 2D Gaussian functions with different $\sigma_x,\sigma_y$ and with an orientation of $\pi/4$. The $\sigma_y$ has a step of $1/3$ while the $\sigma_x=\sigma_y/3$, thus producing an elliptical structure at differen scales. In the rest of Fig. \ref{fig:2d_gauss}, we calculated the scales at the center point of each image with the following configurations. For scale-space, we used the scale-normalized Laplacian.
For shearlets we used the $\mathcal{B}$ measure (Eq. \eqref{eq:blobness}) 
and the 2-nd derivative of the Gaussian was used as $\psi_1$. By analyzing Fig. \ref{fig:2d_gauss}, we can observe that, by performing direct calculations, perfect scale-invariance still holds for the three discussed frameworks on syntectic images.

\graphicspath{ {./images/ch-blobs/log_comparison/} }
\begin{figure}[!t]
    \centering
    \includegraphics[height=0.35\textwidth]{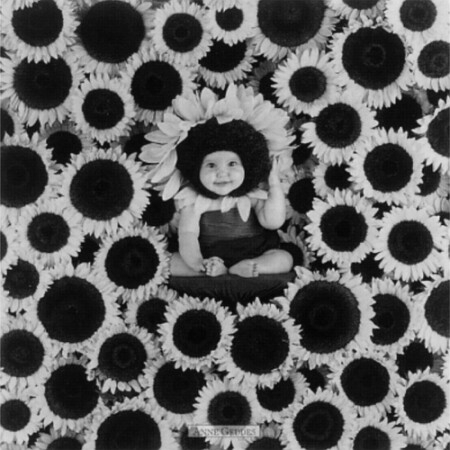}\ \
    \includegraphics[height=0.35\textwidth]{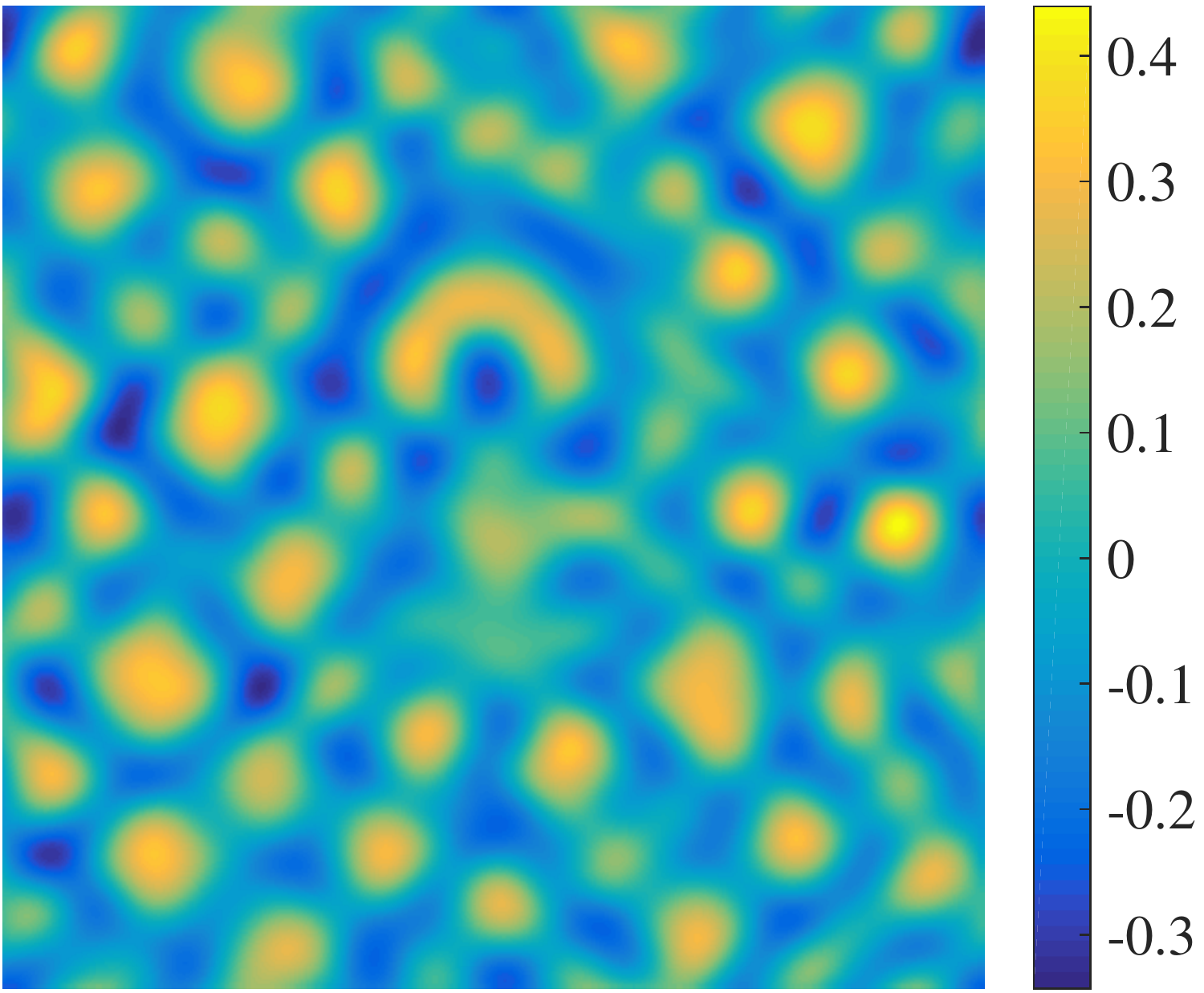}
  \caption{Space localization of blob features using the $\mathcal{B}$ measures at $j=0$.}
  \label{fig:blobs_space}
\end{figure}

\section{Blob Detection with Shearlets}
\label{sec:sbd}

In this section we deal with the problem of automatically detecting blobs and  describe our {\em Shearlet Blob Detector (SBD)} algorithm.

\begin{figure}[!t]
  \centering
  \raisebox{0.125\height}{
  \includegraphics[height=0.35\textwidth]{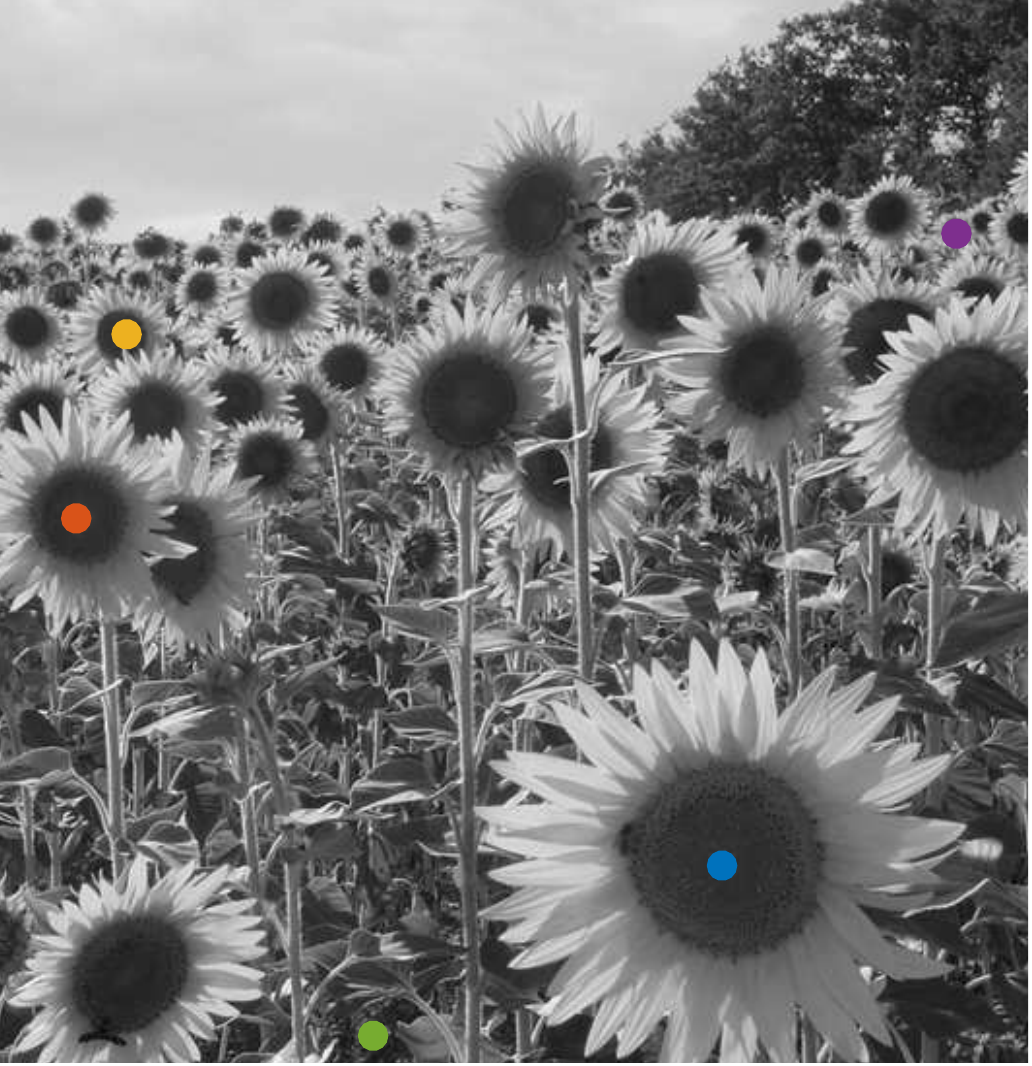}}\ \
  \includegraphics[height=0.4\textwidth]{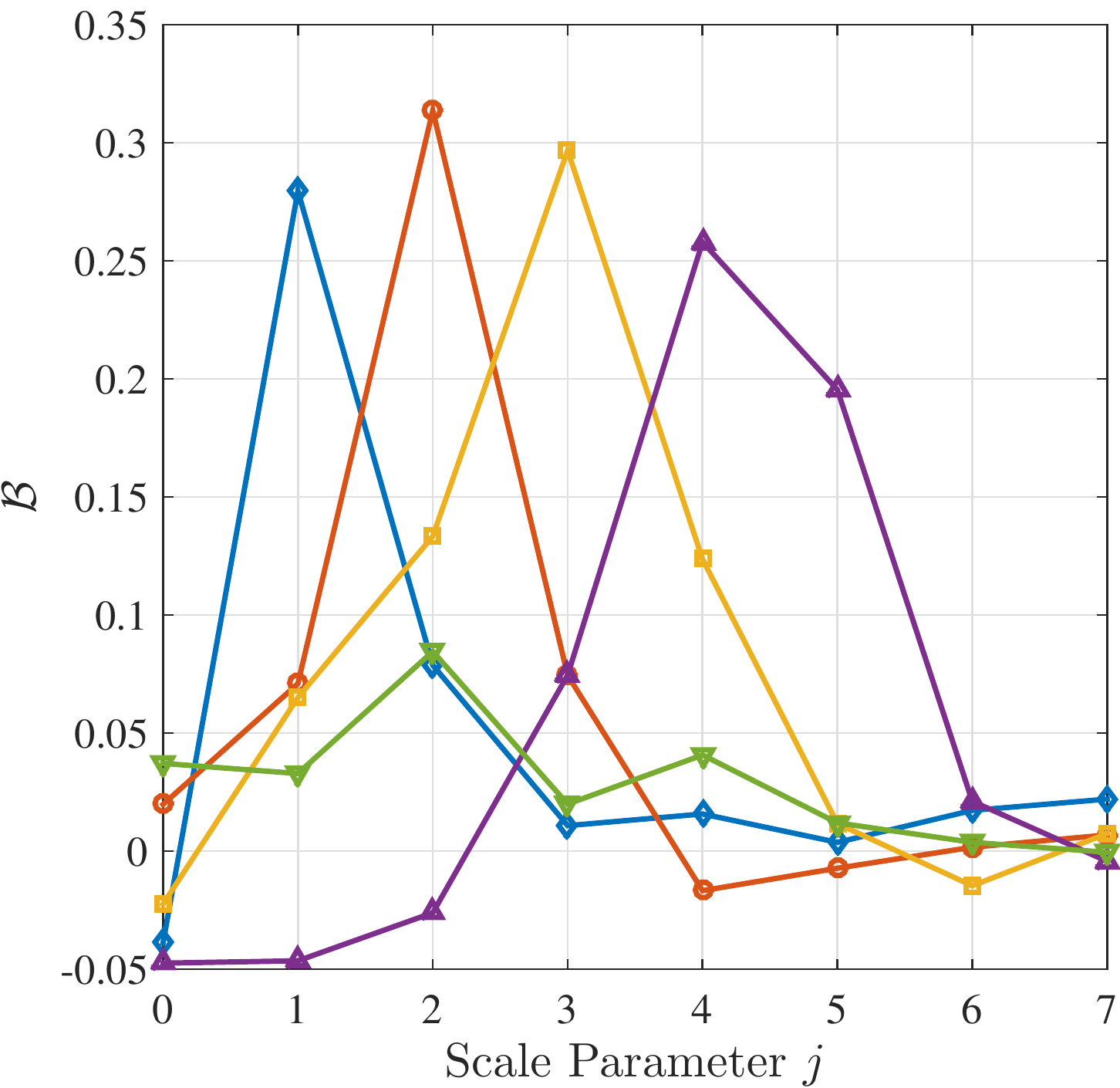}
  \caption{The behavior of ${\cal B}$ across scales for different points, coherently color-coded.}
  \label{fig:features_scales}
\end{figure}

Let us first analyze separately the behavior of $\mathcal{B}$ in scale and space. Fig. \ref{fig:blobs_space} provides the result of the measure ${\cal B}$ computed at scale $j=0$. As we can observe, the biggest sunflower structures are correctly localized by obtaining a high value of $\mathcal{B}$ at the coarsest scale.
Fig. \ref{fig:features_scales} illustrates instead the behavior of ${\cal B}$ across scales for different key points of a real image. We consider in particular five locations corresponding to blob structures of different size and one texturized region. It is easy to observe that although there is no perfect scale invariance, the peaks are clearly visible and their position reflect the different spatial extents of the corresponding image structures.

\begin{figure*}[!t]
 \centering
 \includegraphics[height=0.21\textwidth]{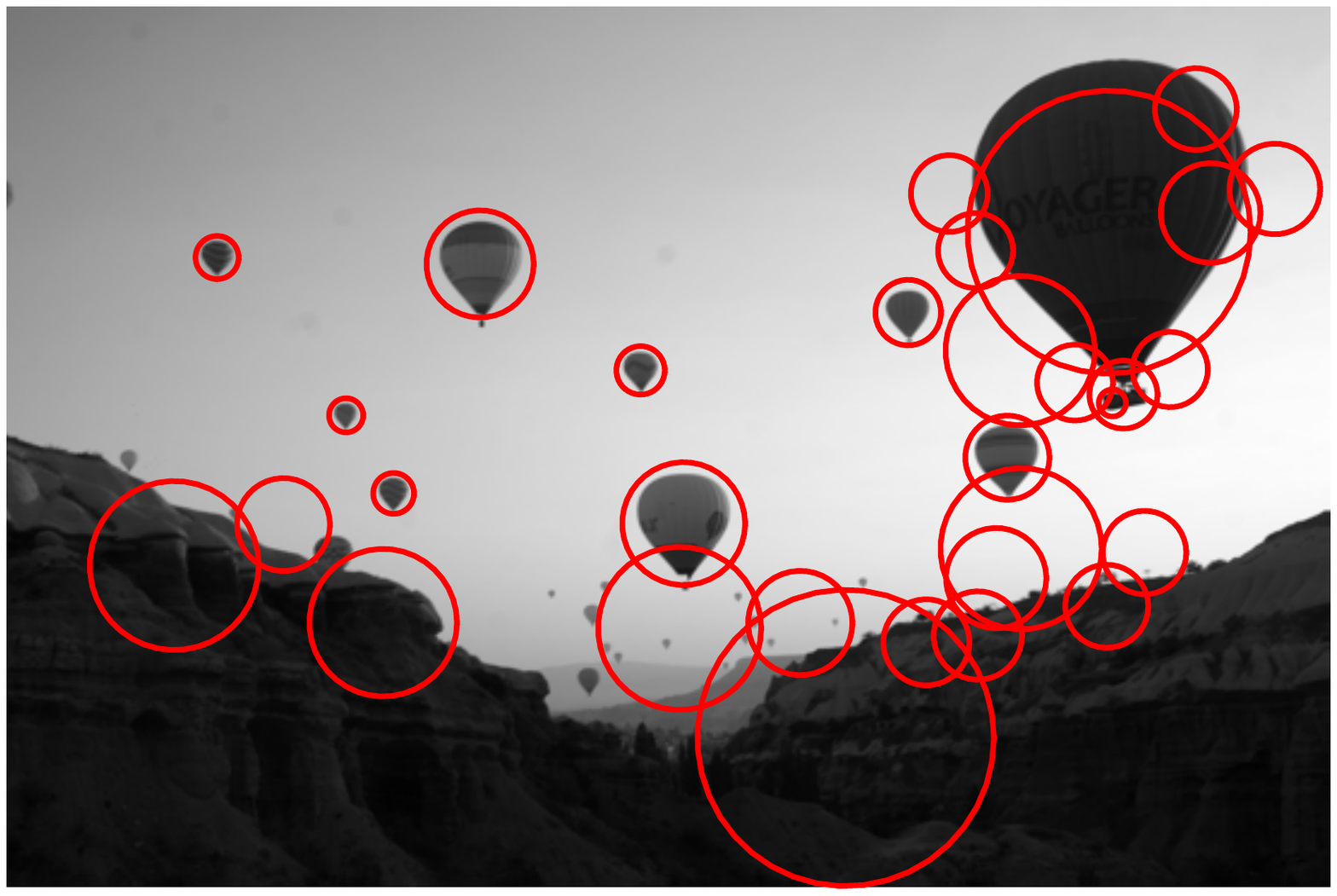}
 \includegraphics[height=0.21\textwidth]{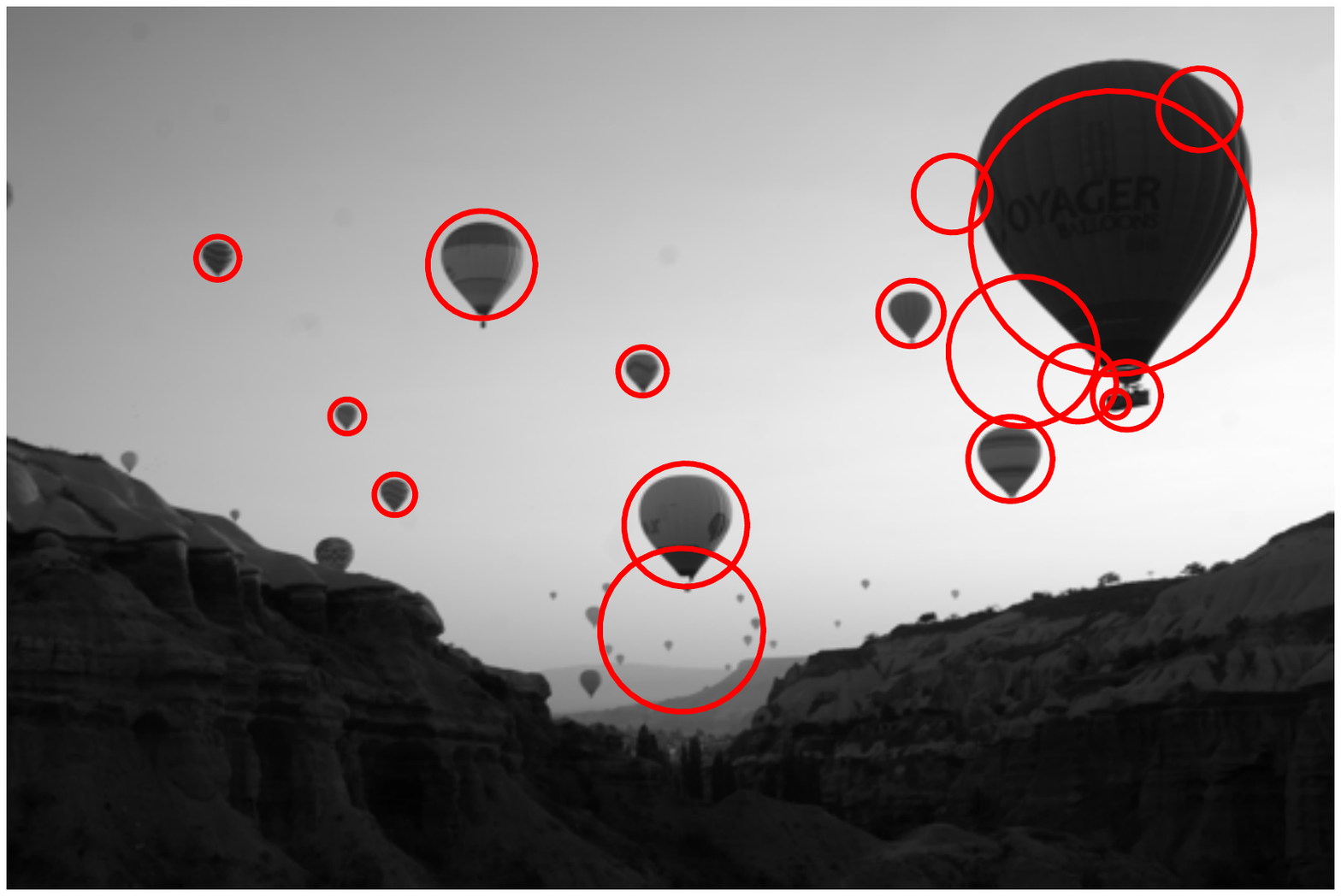}
 \includegraphics[height=0.21\textwidth]{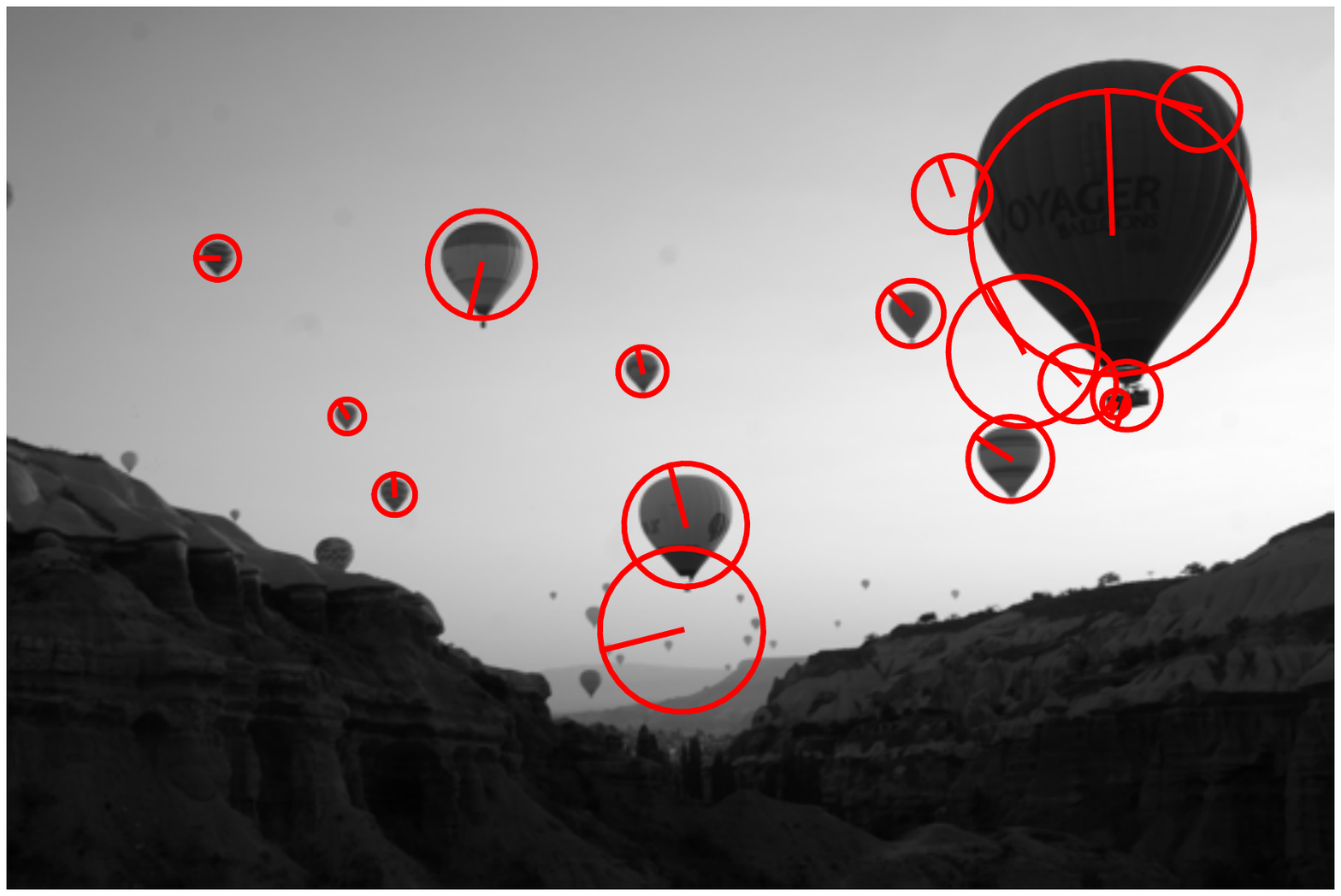}
 \caption{The different steps of the shearlet blob detector algorithm. Left: feature point localization; Centre: edge response elimination; Right: orientation assignment}
\label{fig:sbd_steps}
\end{figure*}

Similarly to the method proposed by Lowe to extract DoG features \cite{lowe2004distinctive}, our approach consists of different steps of measures, computation and refinement. In the reminder of the section we detail the three steps of the blob detection algorithm. Fig. \ref{fig:sbd_steps} shows the intermediate steps results on a sample image.

\graphicspath{{./images/ch-blobs/B_analysis/}}

\subsection{Accurate feature point localization}

A location $m$ at a certain scale $j$ is recognized as a candidate keypoint if the function $\mathcal{B}(m,j)$, computed over a spatial $3\times3\times3$ (2D space $\times$ scales) neighborhood centered on $m$ assumes a local {extrema (maxima or minima)} in it and its value is above a threshold.
\begin{equation}
(\bar{m},\bar{j}) = \arg \underset{m,j}{\operatorname{\mathrm{maxmin}}}\ \mathrm{local}\ \mathcal{B}(m,j).
\end{equation}

Then, the local extrema of the $\mathcal{B}$ function are interpolated in space and scale with the Brown and Lowe method \cite{brown2002invariant} to reduce the effect of considering a limited number of scales. The outcome of this step is a set of feature points with their associated scales.

Notice that since the $\mathcal{B}$ measure is isotropic, only rotationally invariant features will be detected.

\graphicspath{{./images/ch-blobs/sbd_steps/}}

\graphicspath{{./images/ch-blobs/sbd_qualitative_results/}}
\begin{figure*}[!t]
 \centering
 \includegraphics[height=0.235\textwidth]{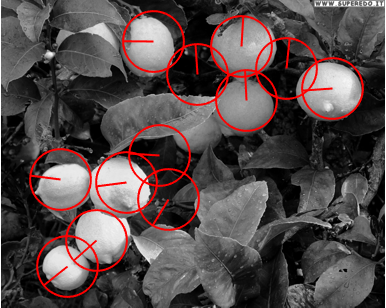}
 \includegraphics[height=0.235\textwidth]{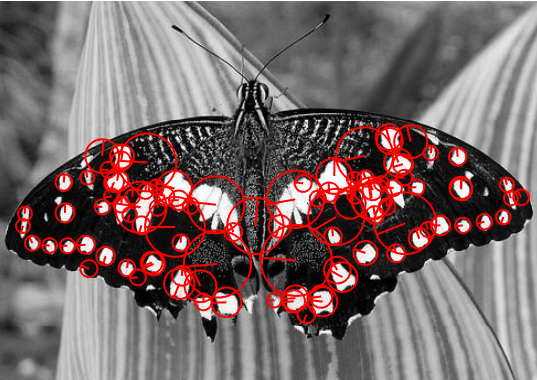}
 \includegraphics[height=0.235\textwidth]{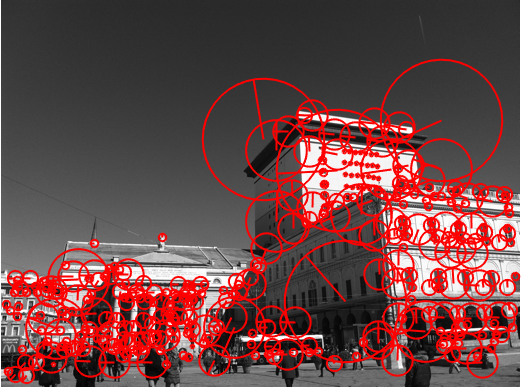}
 \caption{Qualitative results of our blob detector on different scenarios.}
 \label{fig:sbd_qualitative_results}
\end{figure*}

\subsection{Edge responses elimination}
The function $\mathcal{B}$  has strong responses along edges, especially at fine scales. Therefore, in order to increase stability of the detected points, we need to eliminate the feature points that have high edge responses. By using the shearlet transform, we define $\varepsilon_j(m)$ as a metric that measures for a point $m$ at scale $j$ how spread out are the orientation responses with respect to the predominant orientation response. Formally,
\begin{equation*}
    \varepsilon_j(m)=
    \frac{1}{4\lfloor 2^{j/2} \rfloor} \abs{ \sum_k \prt{\mathcal{SH}(\mathcal{I})(j,k,m) - \mathcal{SH}(\mathcal{I})(j,k_\mathrm{max},m)}^2 }
\end{equation*}
\noindent where ${4\lfloor 2^{j/2} \rfloor}$ is the total number of shearings for scale $j$, and $k_{\max}$ is the shearing with larger shearlet response,
\begin{equation}
\label{eq:max_shear}
  k_\mathrm{max} = \arg\max_k |{\ST}(\mathcal{I})(j,k,m)|.
\end{equation}
High values of $\varepsilon_j(m)$ correspond to situations where $k_\mathrm{max}$ is the only orientation with strong support, that is, $m$ is an edge or close to an edge point. Conversely, the value of $\varepsilon_j(m)$ starts decreasing when the point $m$ has more than one, or none, predominant orientations. The second case corresponds to blob points.

Therefore, points $m$ with a high edge response may be deleted by an appropriate thresholding.

\subsection{Accurate orientation assignment}

In this step an orientation is assigned to each feature point. This is an important step with a view to the computation of rotation invariant local feature descriptors. By means of the shearlet transform, the predominant orientation at a point $m$ and scale $j$ is easily obtained by finding the index $k_\mathrm{max}$ given by Equation ~\eqref{eq:max_shear}. However, the orientation estimation at coarse scales may have low accuracy since for $j$ small a few shearings are employed. The effects can be attenuated by finding the extremum of an interpolated parabola for the following three points:
\begin{align*}
& [\theta_{k_\mathrm{max}-1},{\ST}(\mathcal{I})(j,k_\mathrm{max}-1,m)]\\
& [\theta_{k_\mathrm{max}}, {\ST}(\mathcal{I})(j,k_\mathrm{max},m)]\\
& [\theta_{k_\mathrm{max}+1}, {\ST}(\mathcal{I})(j,k_\mathrm{max}+1,m)],
\end{align*}
where $\theta_k$ is the angle associated with the shearing $k$, as in Eq. \eqref{eq:theta_k}.\\

Fig. \ref{fig:sbd_qualitative_results} reports outputs examples    of our blob detector on a variety of scenarios.
For all images, blobs have been detected using a shearlet transform with 8 {scales}, and a rather permissive threshold for edge points elimination.
The circles indicating the presence of blobs have a radius proportional to the estimated optimal scale. As observed, such estimates are very close to the effective spatial extent of image structures.

\section{Feature Description with Shearlets}
\label{sec:shear_desc}

In this section we propose a local feature descriptor based on the shearlet transform, the Shearlet  Local Description algorithm (SLD).

The idea behind our descriptor is to encode the shearlet coefficients computed from the SBD, and thus complete the full detection-description pipeline with a single main computation, the shearlet transform in this case.

Given a feature $F = (m_1, m_2, j, \theta)$, where $m=(m_1,m_2)$ is its location, $j$ its estimated scale and $\theta$ its predominant orientation, our descriptor encodes the shearlet coefficients information from a square region centered on $(m_1, m_2)$, scaled with respect to $j$ and rotated according to $\theta$ (Fig. \ref{fig:sld_sampling}).

\subsection{Shearings with common orientations}

\begin{figure}[!t]
 \centering
 \includegraphics[height=0.45\textwidth]{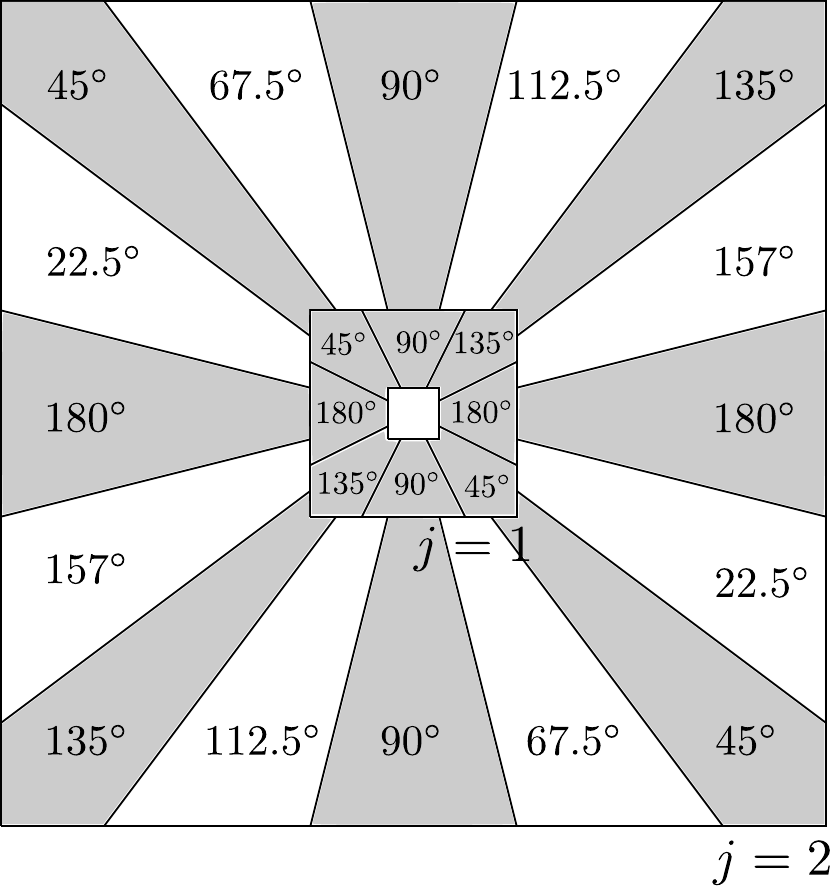}
 \caption{Visualization of the orientations at scales   $j=1,2$ and $O =\{180^\circ, 135^\circ, 90^\circ, 45\}$. In gray: sequence of common orientations $S^O_j$. In white: orientations $S_j$. (see text). }
 \label{fig:common_shears}
\end{figure}

Since in our representation different scales are associated with different amount of shearings, we first need to fix a common number of shearings across scales, in order to obtain descriptions of equal size from keypoints at different scales. Moving towards finer scales, there is an inclusion relation between the corresponding range of shearings, thus the orientations associated to the shearings at coarser scales are also available at fine scales, see Fig. \ref{fig:common_shears}.

Let $O$ be a set of orientations that can be associated with the shearings of {\em all} the employed scales. The cardinality of the set is $|O|=c$, where $c$ is the only parameter\footnote{Notice that $c$ must be a power of 2 to be coherent with the number of shearlets on a scale.} of this method and will influence the size of the descriptor.
By default, $O =\{180^\circ, 135^\circ, 90^\circ, 45\}$ since these are the orientations associated with the four shearlets of the coarser scales $j=0,1$.

We refer to $S_j$ as the sequence of all $4\lfloor 2^{j/2} \rfloor$ shearings for scale $j$, and to $S^O_j$ as the sequence of shearings in scale $j$ with respect to the common orientations $O$. Fig. \ref{fig:common_shears} shows an example. For scale $j=2$, $S_2$ is composed by all the shearlets (white and grey), while $S^O_2$ is only composed by the gray shearlets. Notice that for $j=1$, $S^O_1$ is equal to $S_1$.

\graphicspath{{./images/ch-blobs/sld_sampling/}}

\subsection{Spatial sampling}

We sample a regular grid of 24 points per side around $(m_1, m_2)$ with a sampling step of $s=2^{j_0-j}$, i.e. the inverse shearlet continuous scale of the feature, covering a length of $24s$.

\begin{figure}[!t]
 \centering
 \includegraphics[height=0.45\textwidth]{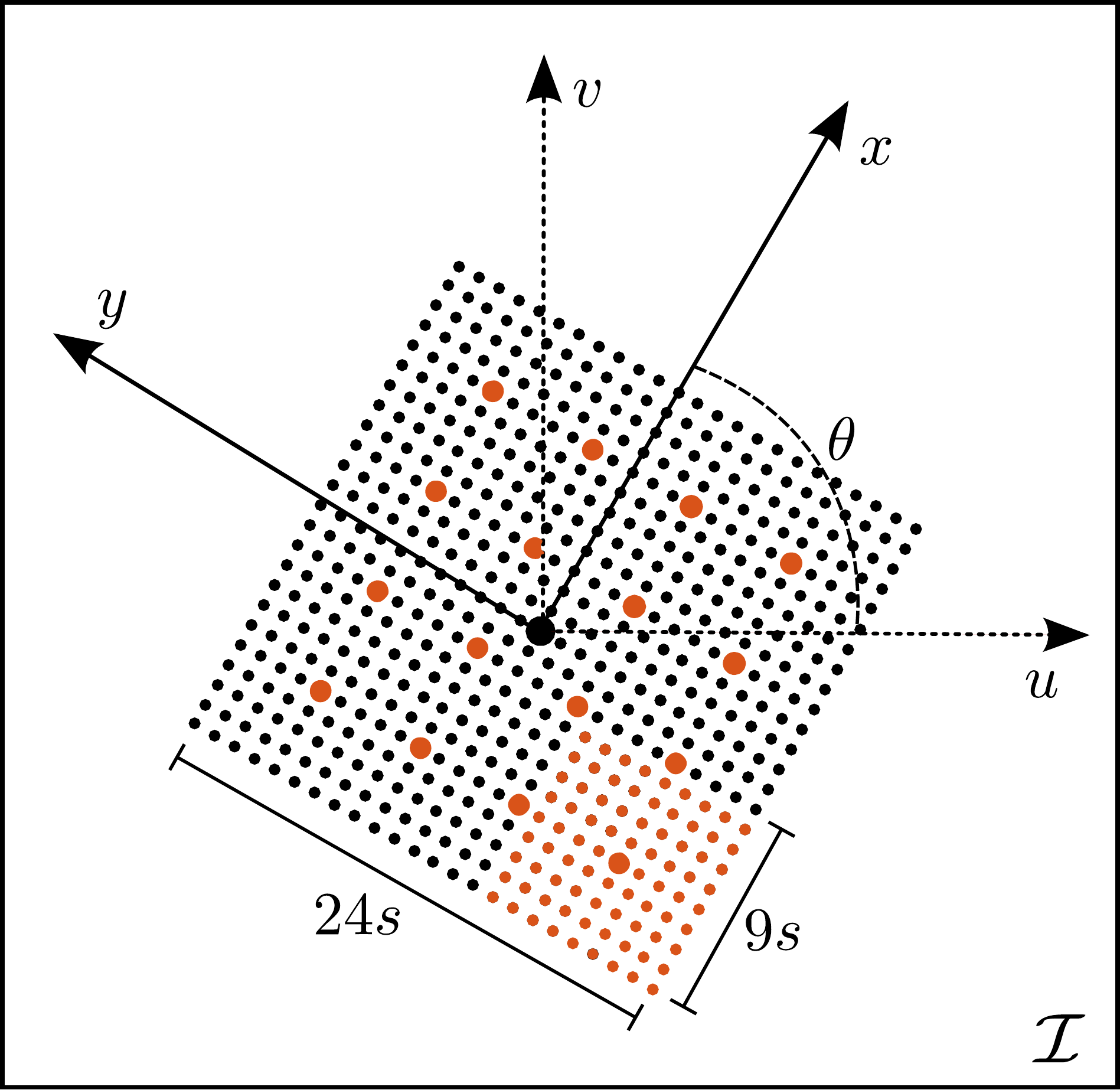}
 \caption{Illustration of the change of coordinates for rotation invariance and the scaled and orientated shearlet sampled points used for the construction of the SLD descriptor.}
 \label{fig:sld_sampling}
\end{figure}

We divide the regular grid in $16$ overlapped subregions of size $9s \times 9s$ (hence including 81 shearlet coefficients). We refer to each subregion using the centroid (thick red points in Fig. \ref{fig:sld_sampling} (b)), which may be described by its relative position with respect to the local keypoint reference system. More formally, the subregions can be referred to as $\{ \mathcal{G}_{e,f}\}$ where $e,f \in \{\pm1,\pm2\}$. Notice that the overlap allows us to cope with small spatial keypoint shifts.

\graphicspath{{./images/ch-blobs/sld_sampling/}}

\subsection{Region rotation}

In order to gain rotation invariance, we compute the actual descriptor on a region which is rotated according to the keypoint main orientation $\theta$. To this purpose, we perform a change of reference system (see Fig. \ref{fig:sld_sampling}),
\begin{equation*}
\begin{pmatrix} x \\ y \end{pmatrix} = s\cdot \begin{pmatrix} \cos\theta & -\sin\theta \\ \sin\theta & \cos\theta \end{pmatrix}\begin{pmatrix} u \\ v \end{pmatrix} + \begin{pmatrix} m_1 \\ m_2 \end{pmatrix}.
\end{equation*}
To maintain the rotation invariance, the shearing parameter $k$ also has to be aligned according to $\theta$. To this purpose, we perform a circular shift following the shearing indexing $i_k$ described on Section \ref{sec:shear_indexing},
\begin{equation*}
t(k) = (i_k \pm n_\theta) \  \mathrm{mod} \ |S_j|
\end{equation*}
\noindent where $n_\theta = \lfloor \theta |S_j| / \pi \rceil$ is the number if shifts required to align with respect to $\theta$.

\subsubsection{Descriptor construction}

The SLD descriptor concatenates statistics on shearlets coefficients in each subregion. We start by describing  a subregion $\mathcal{G}_{e,f}$ of scale $j$ and shearing $k$ by a 2-D vector $\mu(e,f,j,k)$
\begin{equation*}
\mu(e,f,j,k) = \begin{pmatrix} \sum_{(u,v)\in \mathcal{G}_{e,f}} M_j(u,v,k)g(u,v,2.5s),\\ \sum_{(u,v)\in \mathcal{G}_{e,f}} \abs{M_j(u,v,k)}g(u,v,2.5s) \end{pmatrix}.
\end{equation*}
\noindent where
\begin{equation*}
M_j(u,v,k) = \mathcal{SH}(\mathcal{I})(j,t(k),(x,y))
\end{equation*}
\noindent and $g$ is a 2D Gaussian filter with $\sigma=2.5s$. Then, within the subregion, we concatenate $\mu(e,f,j,k)$ for each shearing $k \in S^O_j$ following the order induced by the circular shift: let us denote it as $\mu(e,f,j) \in \mathbb{R}^{2 \times  c}$.

\emph{Remark:} The rotation invariance can be improved by performing a weighted sum over the shearing neighbourhoods of the sampled point, instead of sampling directly into the shearlet coefficient. That is,
\begin{equation*}
M_j(u,v,k) = \sum_{r=-|S_j/2|}^{|S_j/2|} \mathcal{SH}(\mathcal{I})(j,t(k+r),(x,y))g(r,\sigma)
\end{equation*}
where $g$ in this case is a 1D Gaussian with $\sigma = |S_j|/5$. This way small misalignments in the orientation can be overcome. However, the drawback is the additional computation.

Next, the contribution of each subregion is weighted using a Gaussian with $\sigma = 1.5$. and then concatenated to build the descriptor $\mu \in \mathbb{R}^{2 \times c \times 16}$ as
\begin{equation*}
\mu = \left[ \mu(e,f,j) g(e,f,1.5)\right]_{e,f \in \{\pm1,\pm2\}}.
\end{equation*}
Both Gaussian weighting increase robustness towards geometric deformations and localization errors \cite{bay2008speeded}.

\subsubsection{Descriptor normalization}

Finally, in order to gain invariance to linear contrast changes, we normalized the descriptor to a unit vector, using the $\ell_2$ normalization,
\begin{equation}
SLD(F) = \mu / \|\mu\|_2.
\end{equation}

\section{Experimental Results}
\label{sec:blob_exp_results}

In this section we provide an experimental assessment of our shearlet-based method for blob detection and description. Our evaluation follows the Mikolajczyk's protocol and image sequences\footnote{\url{http://www.robots.ox.ac.uk/~vgg/research/affine/}} implemented on VLBechmarks \cite{lenc2012vlbenchmarks}. Each sequence includes 6 images of natural textured scenes with increasing geometric and photometric transformations.

\graphicspath{ {./images/experiments/vlbenchmarks_experiments/repeatability/} }
\begin{figure}[p]
  \centering
  \includegraphics[height=11pt]{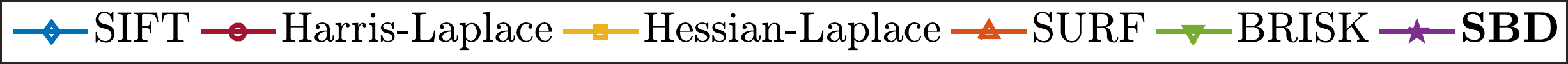}\\
  \subfloat[\label{fig:vprepgraff}Graffiti]{\includegraphics[height=0.45\textwidth]{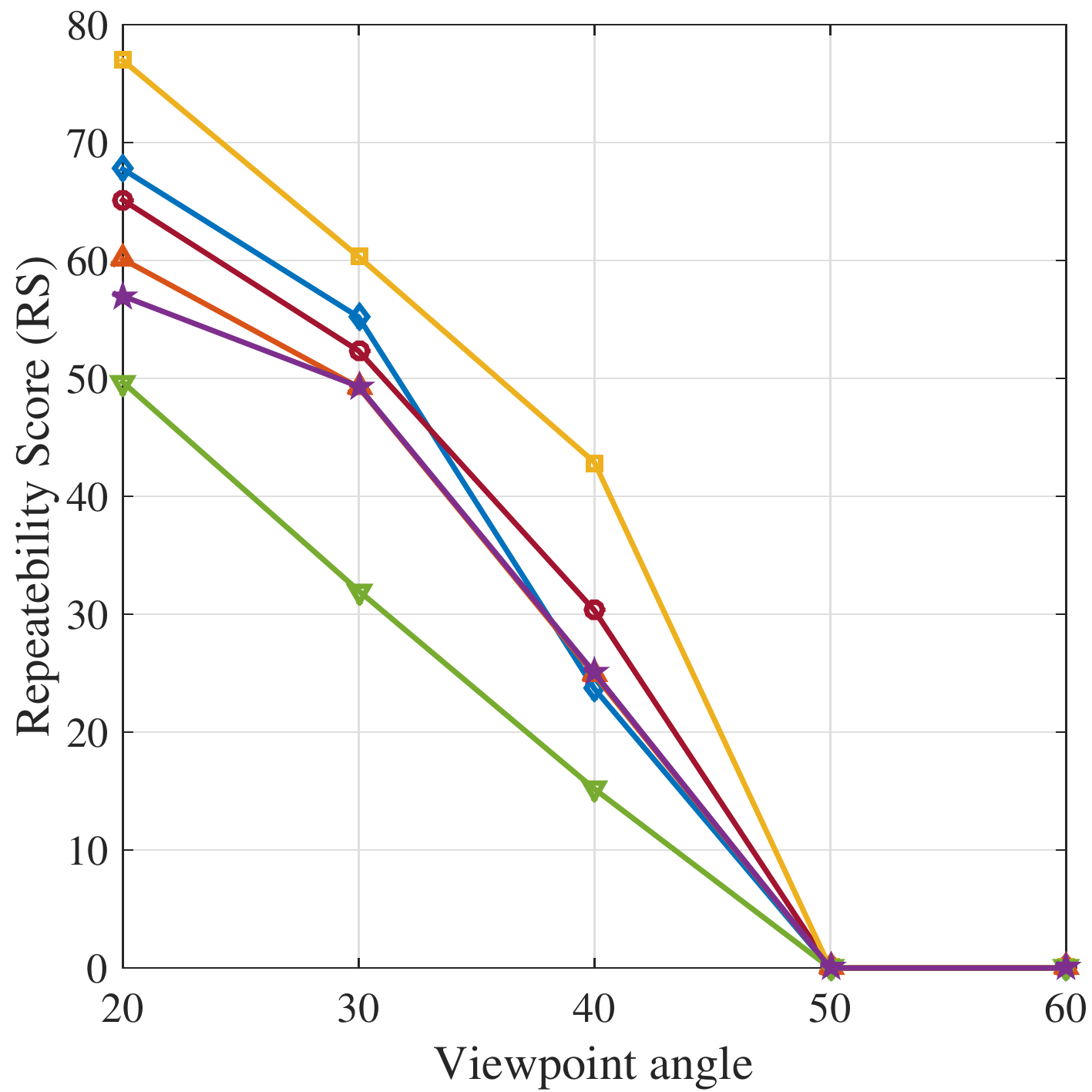}}
  \subfloat[\label{fig:vprepwall}Wall]{\includegraphics[height=0.45\textwidth]{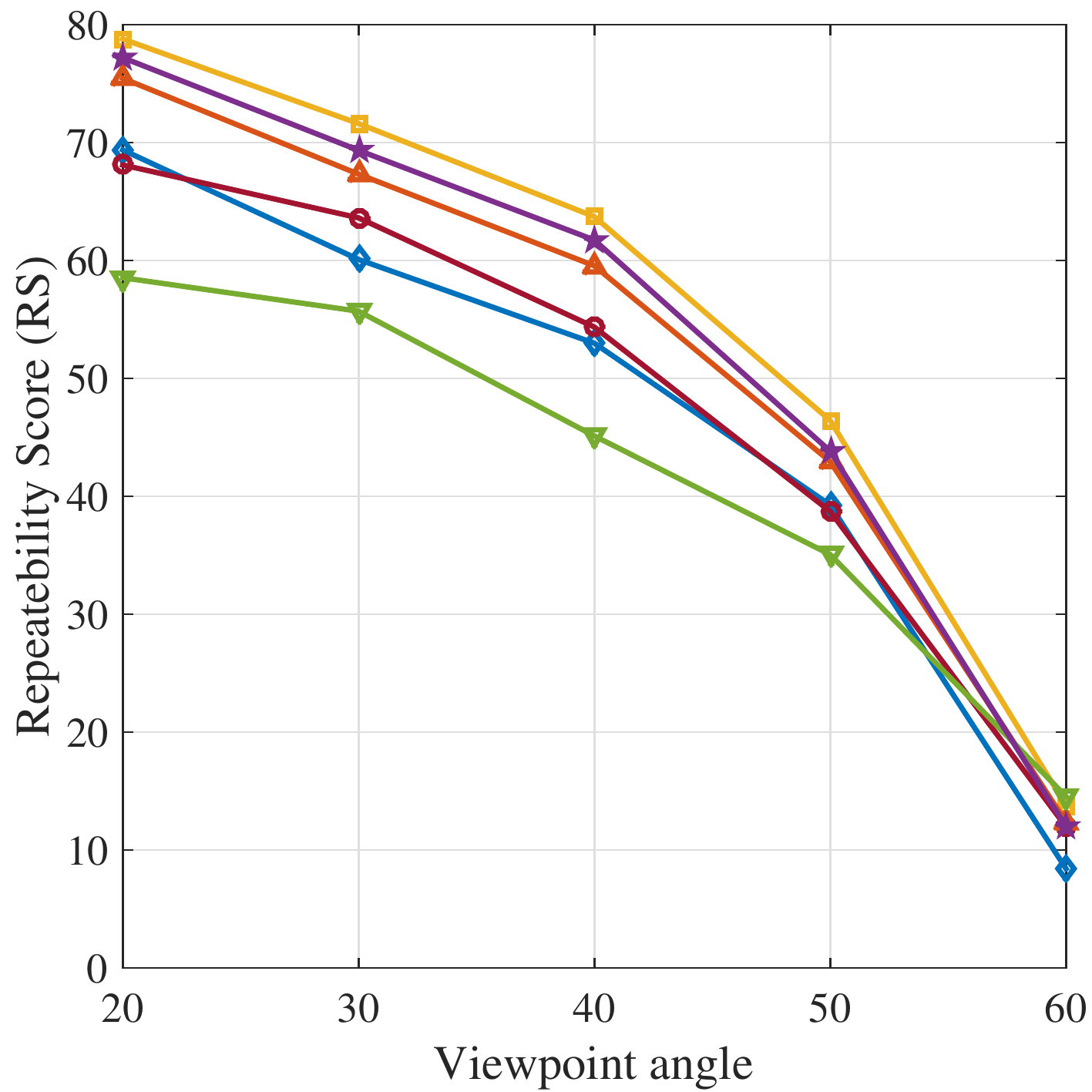}}\\
  \subfloat[\label{fig:screpboat}Boat]{\includegraphics[height=0.45\textwidth]{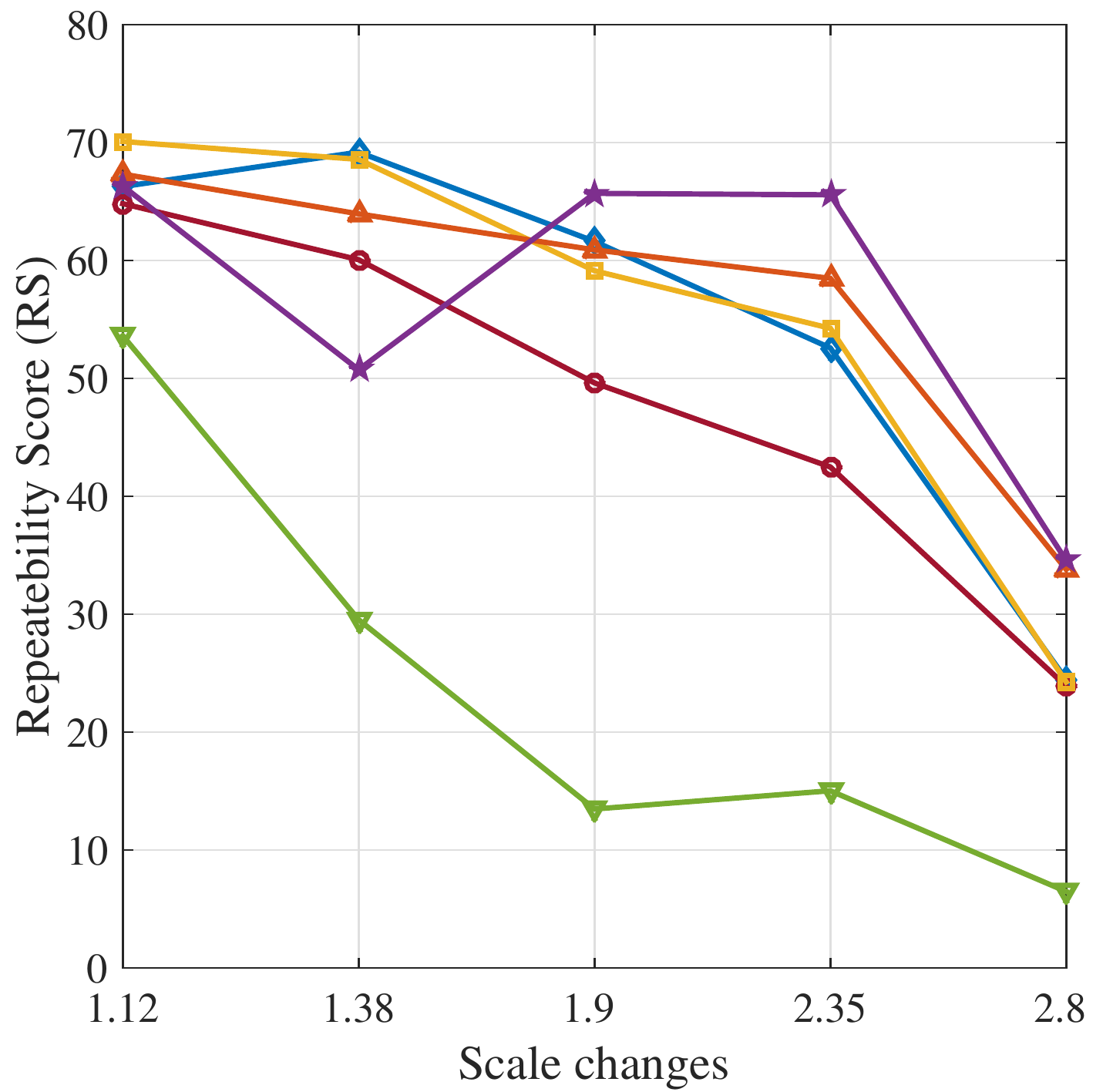}}
  \subfloat[\label{fig:screpbark}Bark]{\includegraphics[height=0.45\textwidth]{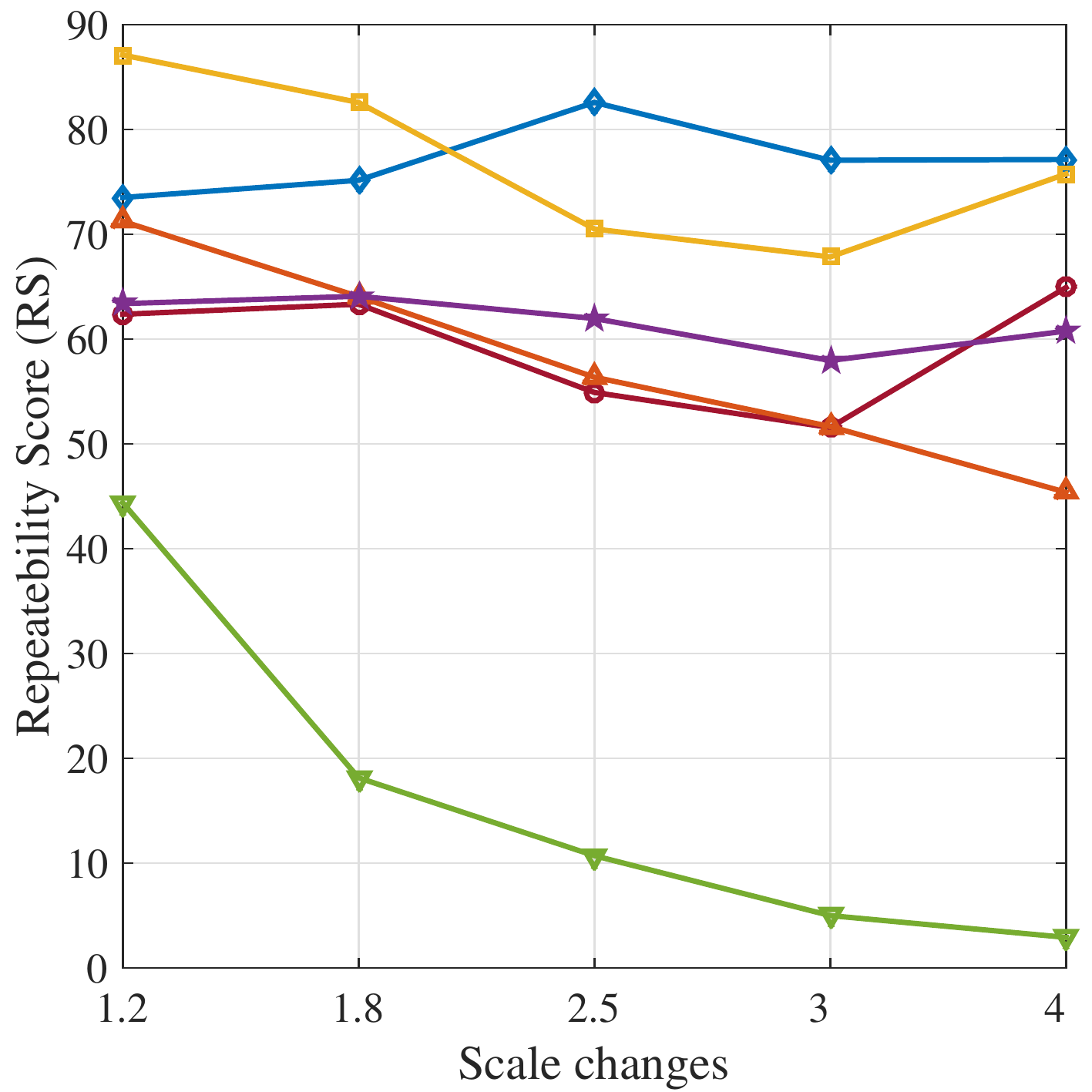}}\\
  \subfloat[\label{fig:blurrepbikes}Bikes]{\includegraphics[height=0.45\textwidth]{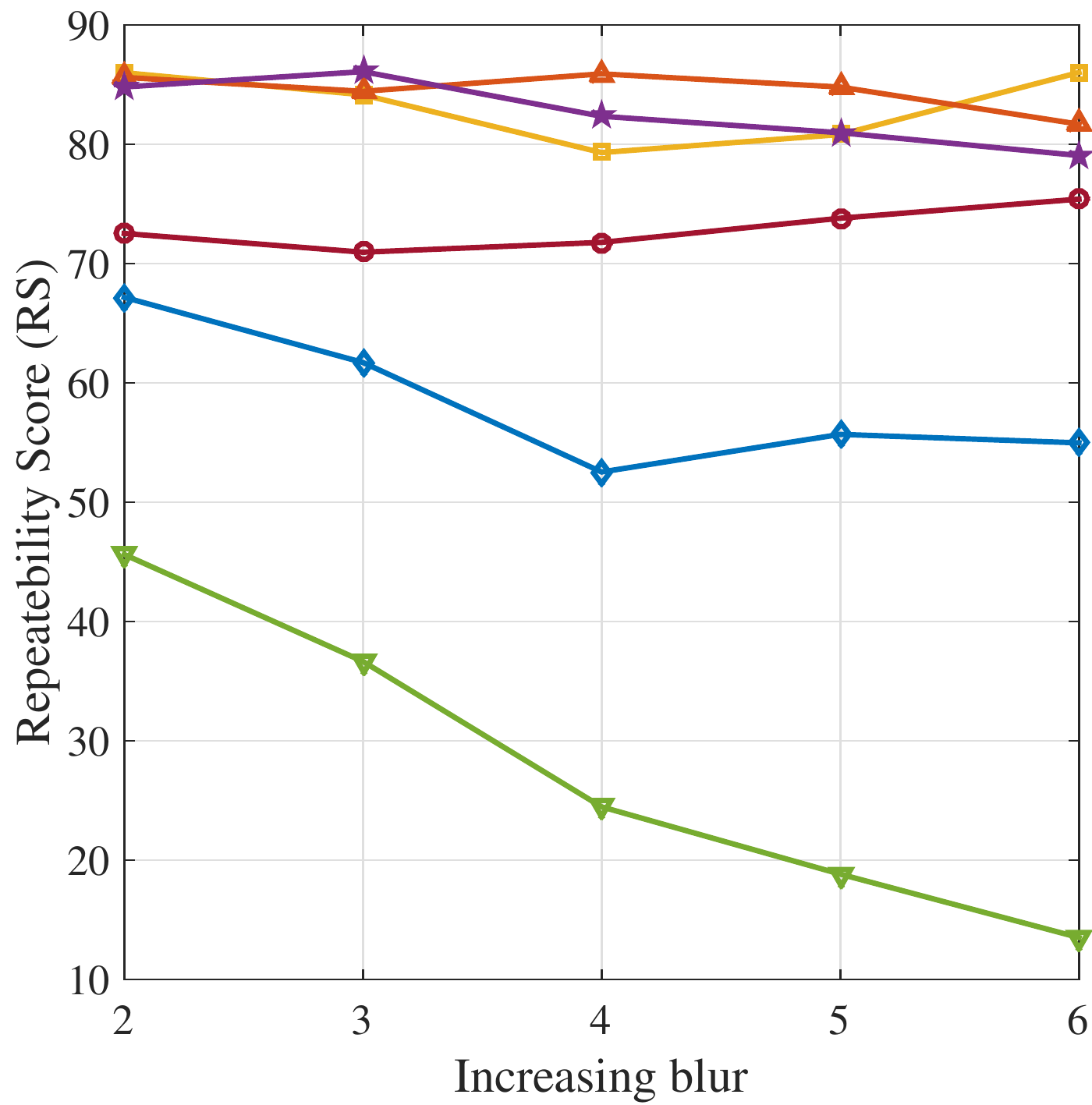}}
  \subfloat[\label{fig:blurreptrees}Trees]{\includegraphics[height=0.45\textwidth]{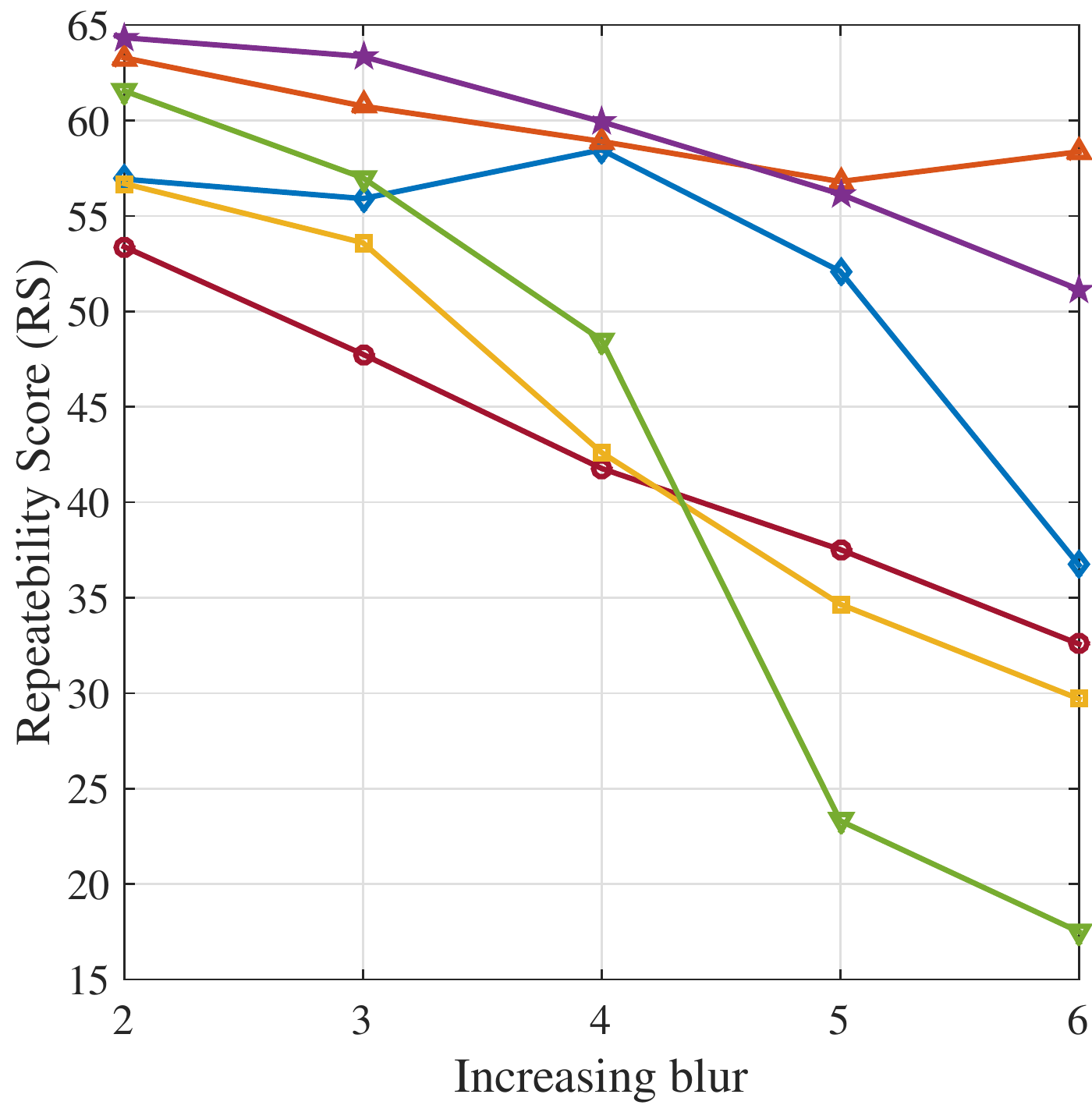}}
  \caption{Comparison of multi-scale feature detection using repeatability scores ($40\%$ overlap error) with respect to viewpoint change (\ref{fig:vprepgraff} and \ref{fig:vprepwall}), scale change (\ref{fig:screpboat} and \ref{fig:screpbark}) and image blur (\ref{fig:blurrepbikes} and \ref{fig:blurreptrees}).}
\label{fig:exp_repeatability}
\end{figure}

\begin{figure}[t]
  \ContinuedFloat
  \centering
  \includegraphics[height=11pt]{legend}\\
  \subfloat[\label{fig:lcrepleuven}Leuven]{\includegraphics[height=0.45\textwidth]{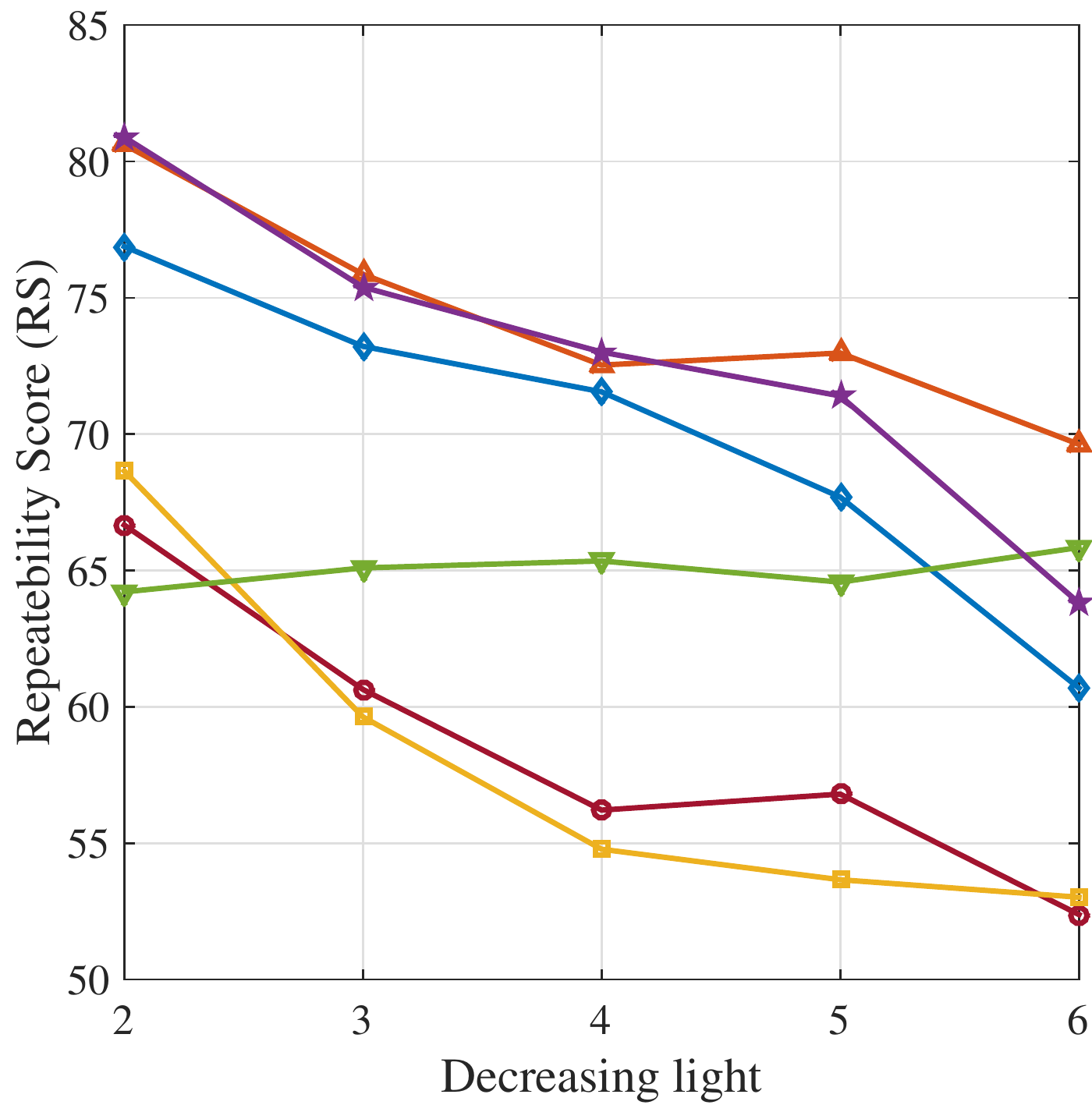}}
  \subfloat[\label{fig:jpegrepubc}Ubc]{\includegraphics[height=0.45\textwidth]{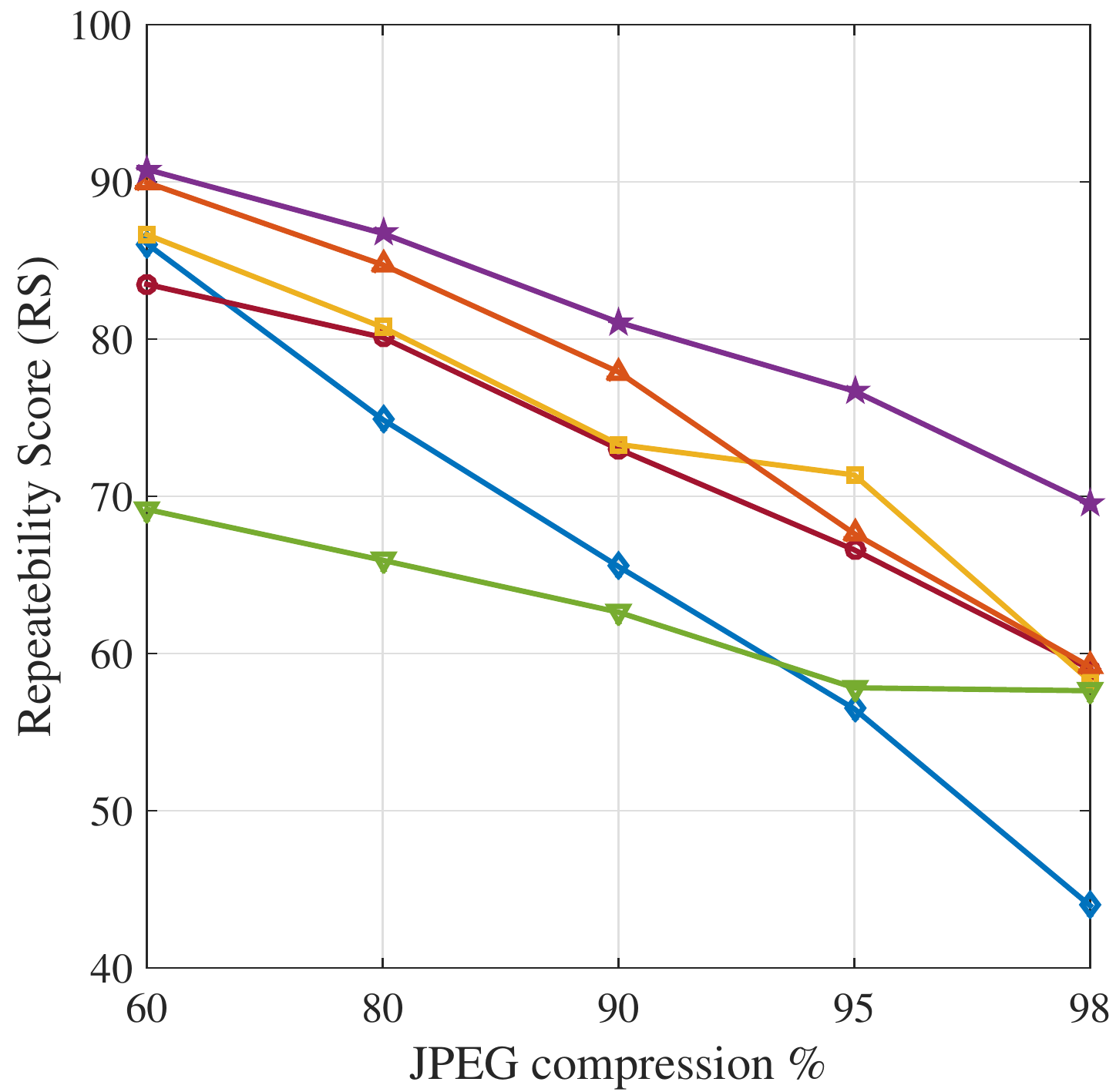}}
  \caption{Comparison of multi-scale feature detection using repeatability scores ($40\%$ overlap error) with respect to light change (\ref{fig:lcrepleuven}) and image compression (\ref{fig:jpegrepubc}).}
\end{figure}

\subsection{Detection Evaluation}

{
We evaluate the detection performances using the \emph{repeatability score} (RS) \cite{mikolajczyk2005comparison}, i.e. the ratio of the number of correspondences and the number of detected features.
We compare the proposed Shearlet Blob Detector (SBD) with SIFT \cite{lowe2004distinctive}, Harris-Laplace \cite{mikolajczyk2004scale}, Hessian-Laplace \cite{mikolajczyk2004scale}, SURF \cite{bay2008speeded} and BRISK \cite{leutenegger2011brisk} feature detectors. As for SIFT, Harris-Laplace and Hessian-Laplace we relied on the implementations provided with the VLBechmarks, while for SURF and BRISK, we adopted the implementation available in MATLAB. For a fair comparison, we adjust the thresholds of the detectors so that the number of detected keypoints is similar in the first image of the sequences. 

{Figure  \ref{fig:exp_repeatability} summarizes the obtained results and shows how our detector is appropriate under general circumstances, while there is no method which is clearly and uniformly superior to the others. }

{\bf Viewpoint changes} (Fig. \ref{fig:vprepgraff} and \ref{fig:vprepwall}). On the Graffiti sequence, Hessian-Laplace outperforms the other methods that show comparative results with exception of BRISK. Hessian-Laplace, SBD and SURF outperforms the others detectors in that order on the Wall sequence.

{\bf Scale changes} (Fig. \ref{fig:screpboat} and \ref{fig:screpbark}). On the Boat sequence, our method allows for higher repeatability as the scale change increases. SIFT and Hessian-Laplace are the best-performing methods on the Bark sequence.

{\bf Image blur} (Fig. \ref{fig:blurrepbikes} and \ref{fig:blurreptrees}). On the Bikes sequence, SURF, SBD and Hessian-Laplace outperform the other detectors with comparable results, while on the Tree sequence SBD outperforms the others detectors with the exception of the last transformation in which SURF obtains better repeatability.

{\bf Illumination changes} (Fig. \ref{fig:lcrepleuven}). On the Leuven sequence, SBD and SURF outperform the rest of detectors. On the first half of the sequence they show very similar repeatability, while for the rest of the transformation SBD is outperformed by SURF.

{\bf JPEG compression} (Fig. \ref{fig:jpegrepubc}). Overall, SBD outperforms the competitors, showing the ability to deal with increasing levels of signal compression.
}

\subsection{Descriptor Evaluation}

\begin{figure}[p]
  \centering
  \subfloat[\label{fig:pr_graff}Graffiti]{\includegraphics[height=0.45\textwidth]{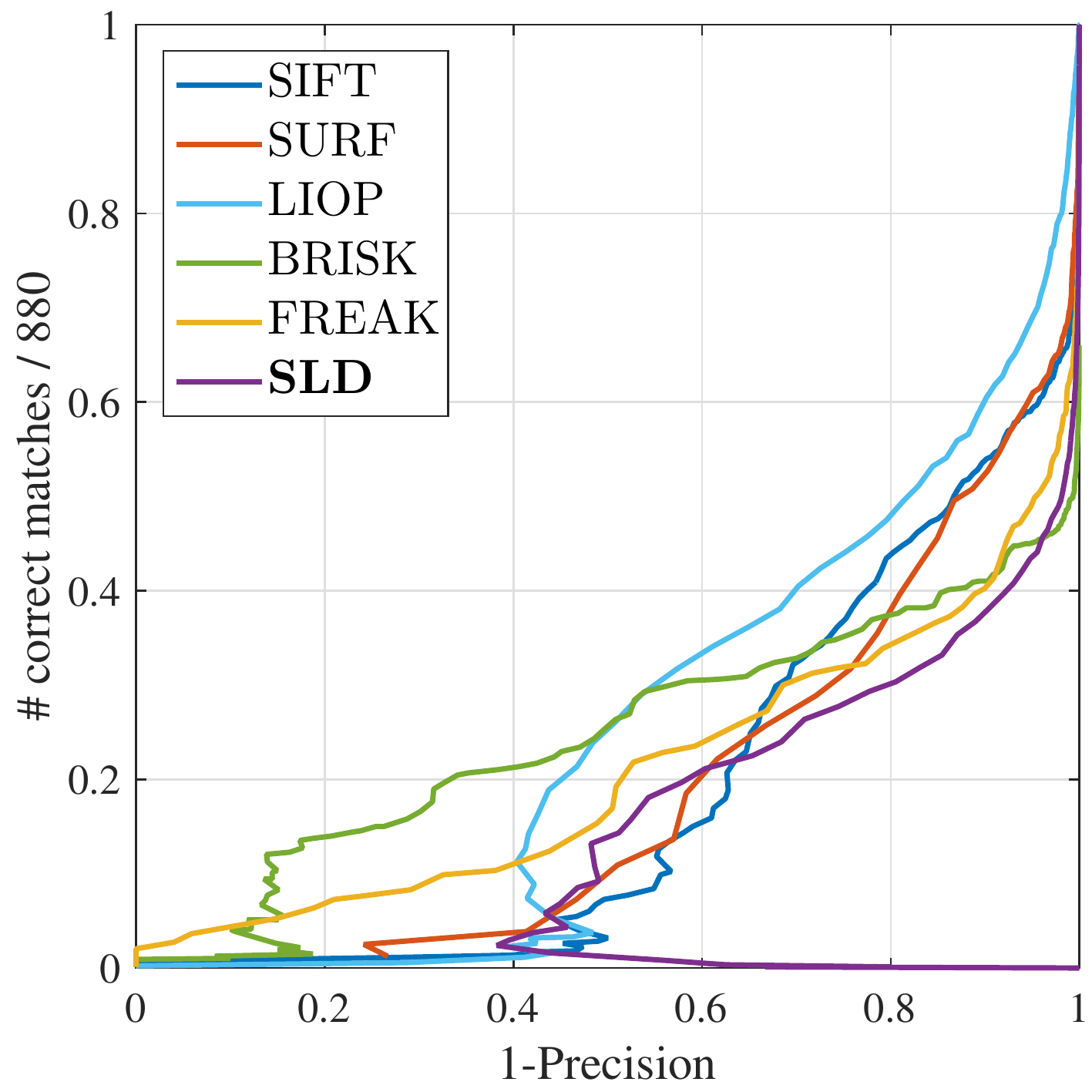}}
  \subfloat[\label{fig:pr_boat}Boat]{\includegraphics[height=0.45\textwidth]{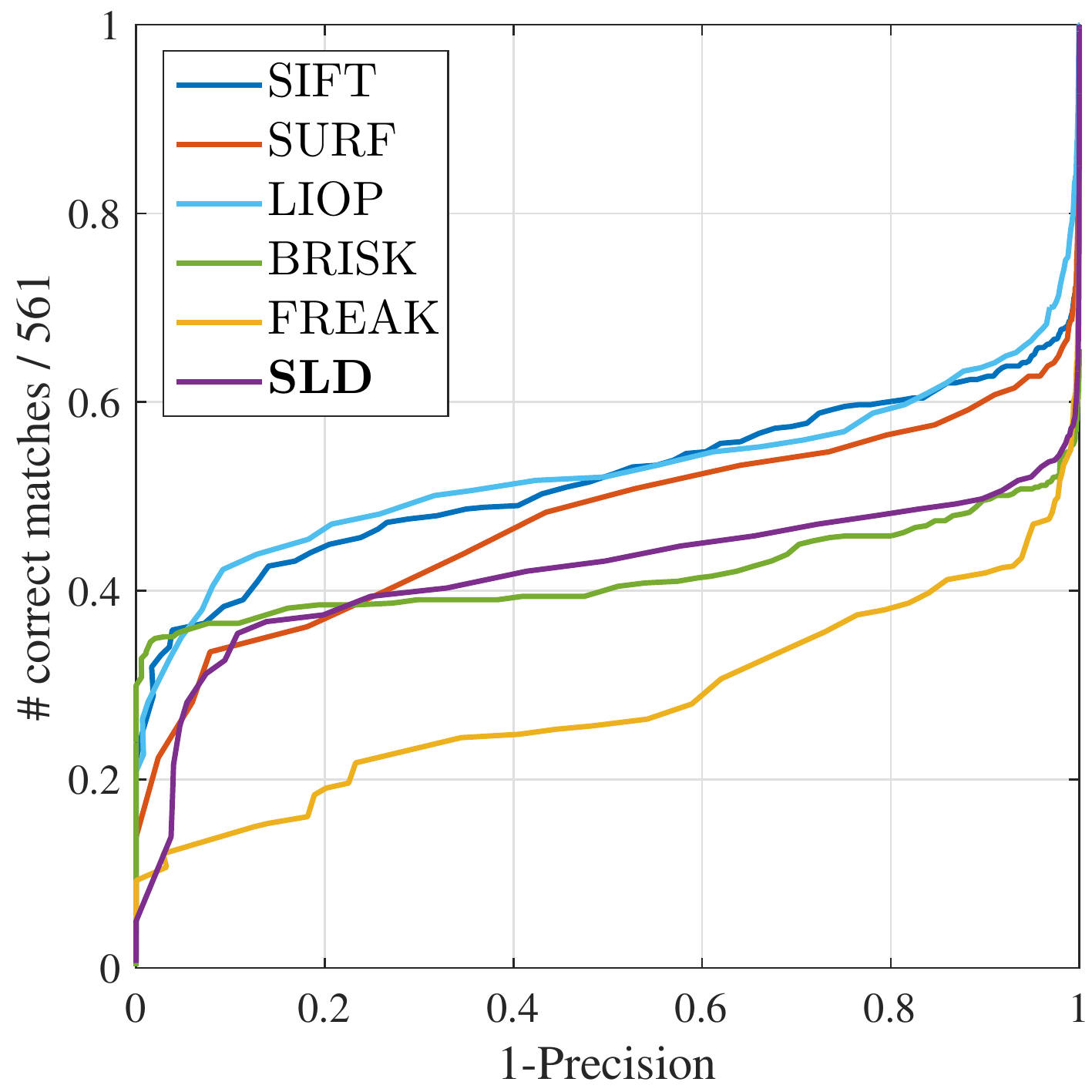}}\\
  \subfloat[\label{fig:pr_bikes}Bikes]{\includegraphics[height=0.45\textwidth]{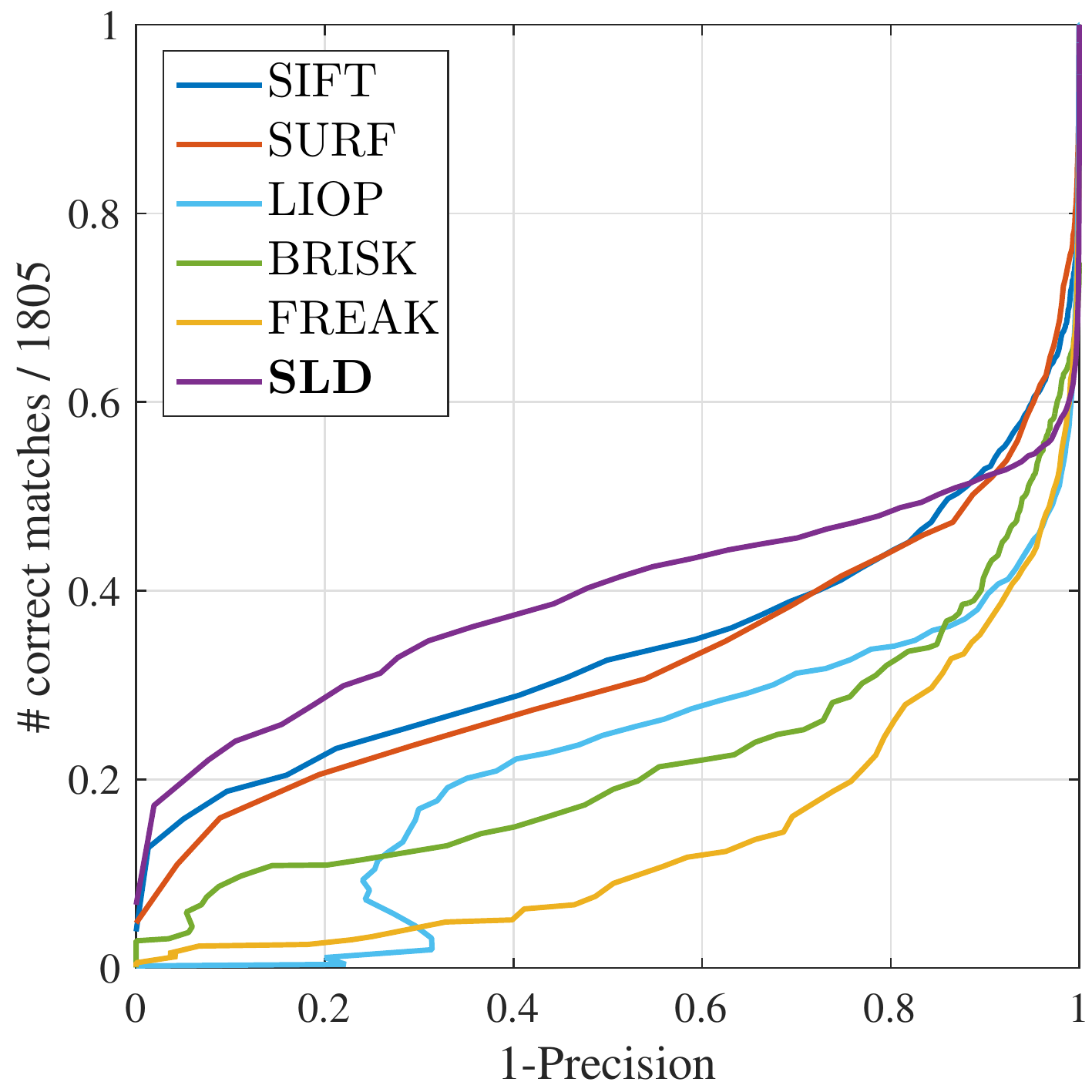}}
  \subfloat[\label{fig:pr_trees}Trees]{\includegraphics[height=0.45\textwidth]{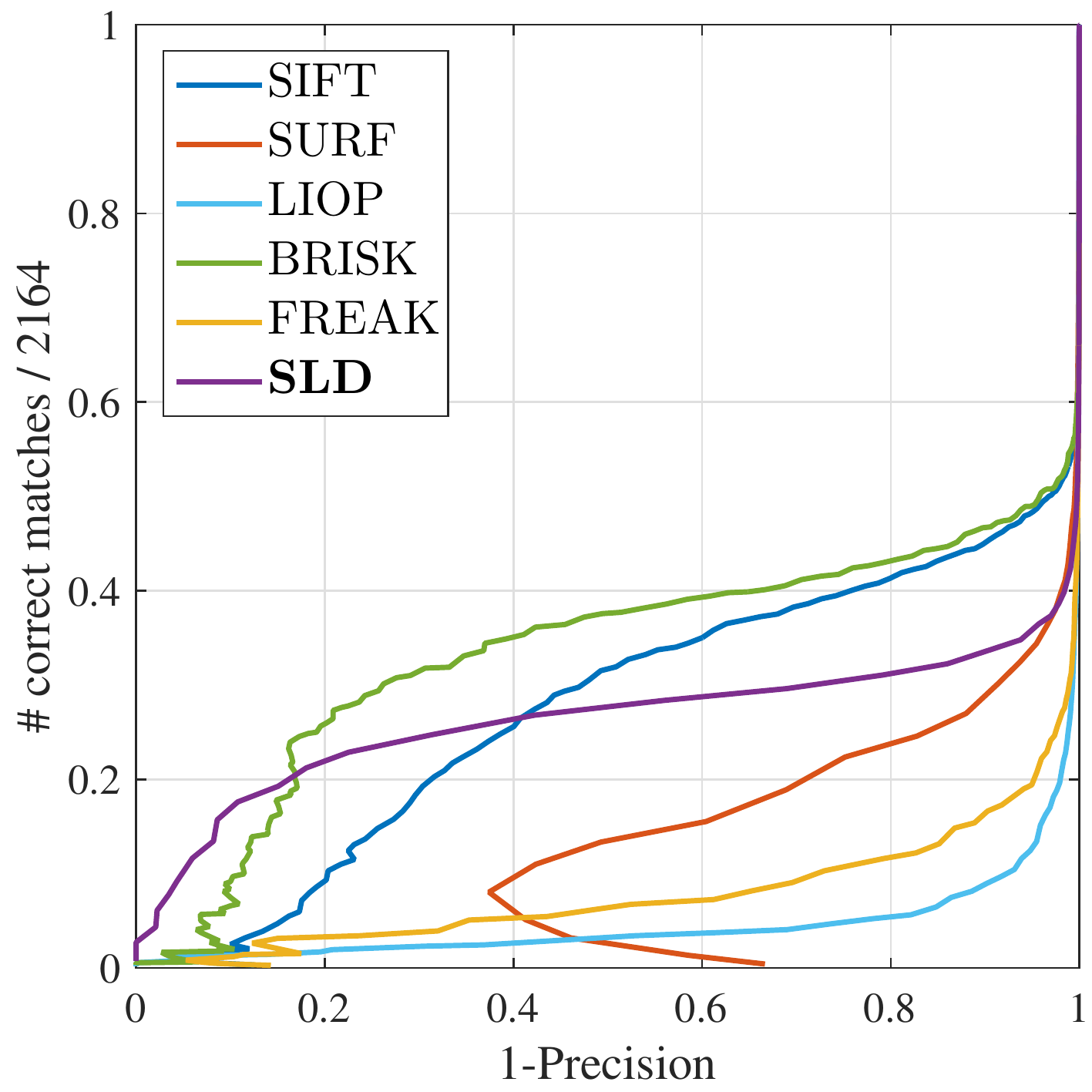}}\\
  \subfloat[\label{fig:pr_leuven}Leuven]{\includegraphics[height=0.45\textwidth]{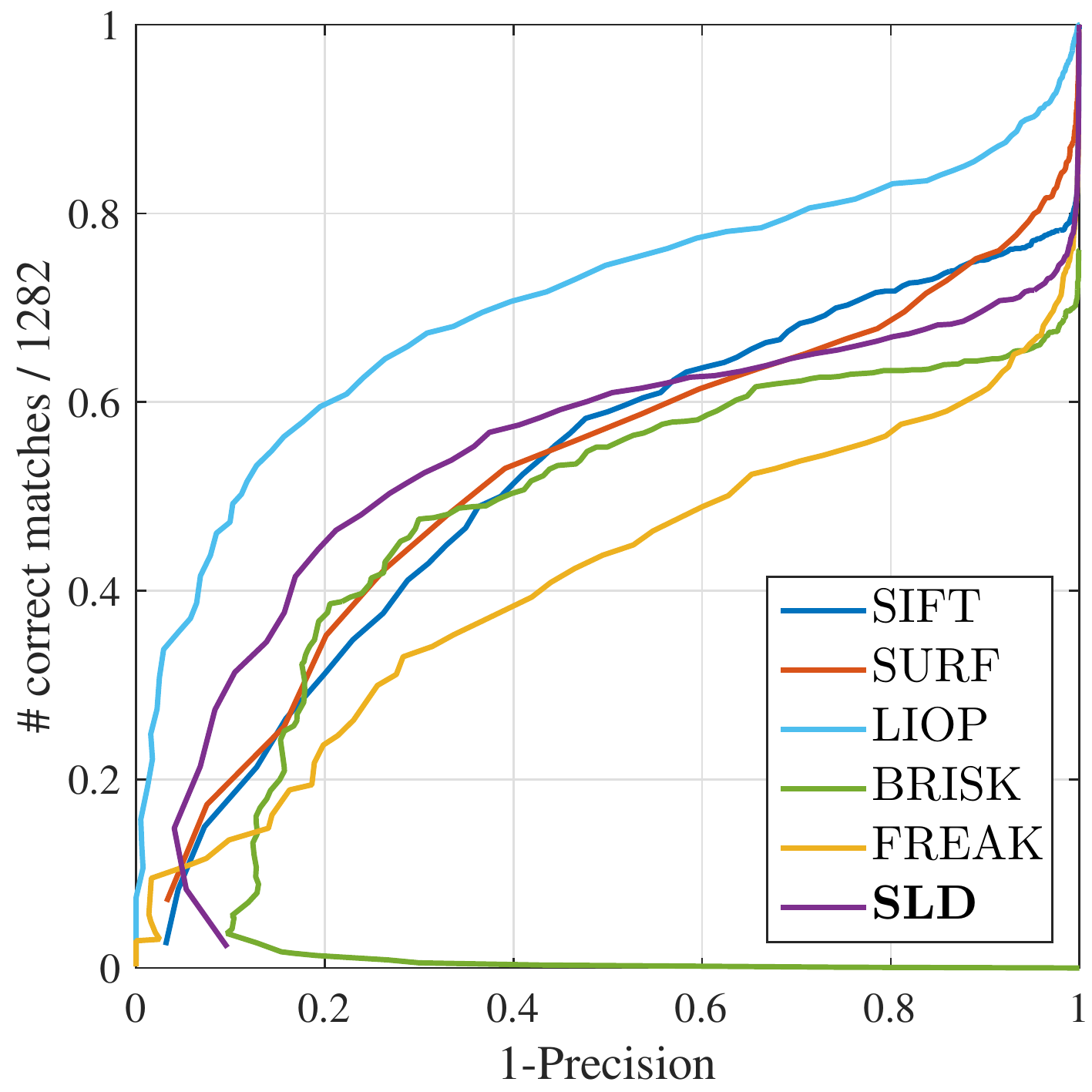}}
  \subfloat[\label{fig:pr_ubc}UBC]{\includegraphics[height=0.45\textwidth]{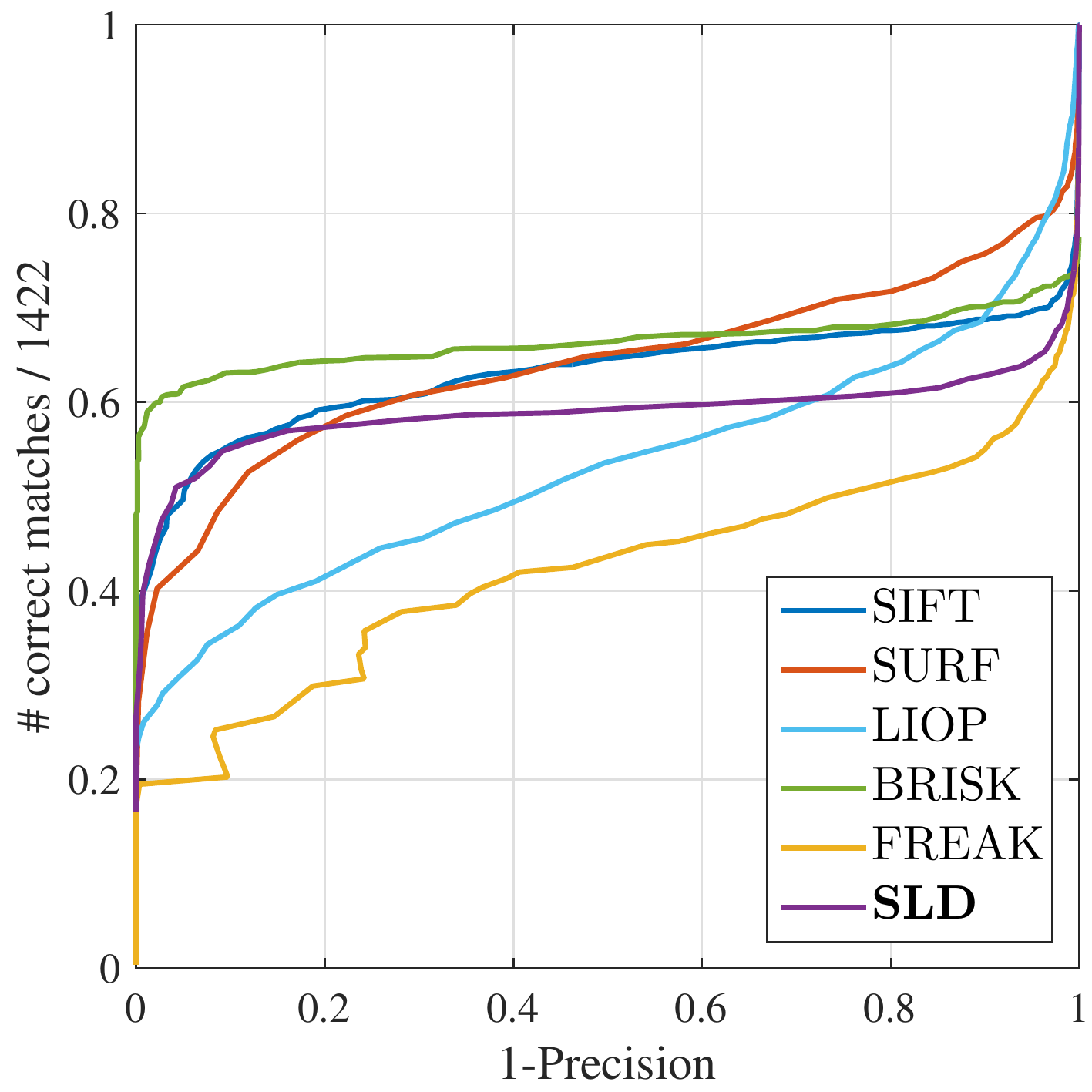}}
  \caption{Comparison of feature descriptors using Precision-Recall curves with respect to viewpoint change (\ref{fig:pr_graff}), scale change(\ref{fig:pr_boat}), image blur (\ref{fig:pr_bikes} and \ref{fig:pr_trees}), light change (\ref{fig:pr_leuven}), and image compression (\ref{fig:pr_ubc}).}
\label{fig:exp_precision_recall}
\end{figure}

{
The Shearlet Local Descriptor (SLD) is evaluated using \emph{recall} (number of correct matches / number of correspondences) vs \emph{1-precision} (number of false matches / number of matches) curves obtained by matching pairs of images ($1^{st}$ and $4^{th}$) from each sequence (as in \cite{mikolajczyk2005performance}). As a comparative evaluation, we also report the results obtained using SIFT, SURF and LIOP \cite{wang2011local} descriptors, along with the recent BRISK and FREAK \cite{alahi2012freak} (implementation available in MATLAB). For all the descriptors we employed their default parameters included our SLD, for which we set $c = 4$. Our default choice represents a compromise between computational efficiency and quality of the results. As for matching strategy, we used a threshold-based approach -- where two detected blobs are matched if the distance between their descriptors is below a threshold -- which is known to be indicative of the distribution of the descriptors in the space \cite{mikolajczyk2005performance}. We employed the Euclidean distance for SLD, SIFT, SURF and LIOP, while we compare BRISK and FREAK (binary descriptors) using the Hamming distance. In order to maintain an unbiased comparison, we evaluate all the descriptors in combination with the DoG feature detector (the one usually coupled with SIFT) using the default parameters for all the image sequences. We now discuss the obtained results, reported in Figure \ref{fig:exp_precision_recall}.

{\bf Viewpoint changes} (Fig. \ref{fig:pr_graff}). On the Graffiti sequence, BRISK and LIOP results as the best performing descriptor.

{\bf Scale changes} (Fig. \ref{fig:pr_boat}). On the Boat sequence, SIFT and LIOP descriptors guarantee the highest accuracies, while our method provides intermediate performances.

{\bf Image blur} (Fig. \ref{fig:pr_bikes} and \ref{fig:pr_trees}). On the Bikes sequence, SLD outperforms the competitors, while SIFT and SURF show consistent results. BRISK is the best performing approach for the Trees sequence, while SLD provides a trade-off between SIFT and SURF results.

{\bf Illumination changes} (Fig. \ref{fig:pr_leuven}). On the Leuven sequence, LIOP is the best performing descriptor, while SLD outperforms the rest of the methods, except for very low precision values.

{\bf JPEG compression} (Fig. \ref{fig:pr_ubc}). On the UBC sequence, SLD, SIFT, SURF and BRISK show overall comparable results, while FREAK and LIOP provides poor performances.
}

{In summary, here again, we do not observe a clear superiority of a method over the others. It should be noticed how our SLD behaves consistently well in the presence of blur, compression effects, illumination changes. This is explicable in terms of the properties of shearlets which provide us with an optimal sparse representation for natural images. The potential of shearlets under these circumstances is further evaluated in the next section.}

\graphicspath{ {./images/experiments/vlbenchmarks_experiments/precision-recall/} }

\graphicspath{ {./images/experiments/copydays_experiments/} }
\begin{figure}[!t]
\centering
    \subfloat[JPEG compression]{
    \includegraphics[height=0.45\textwidth]{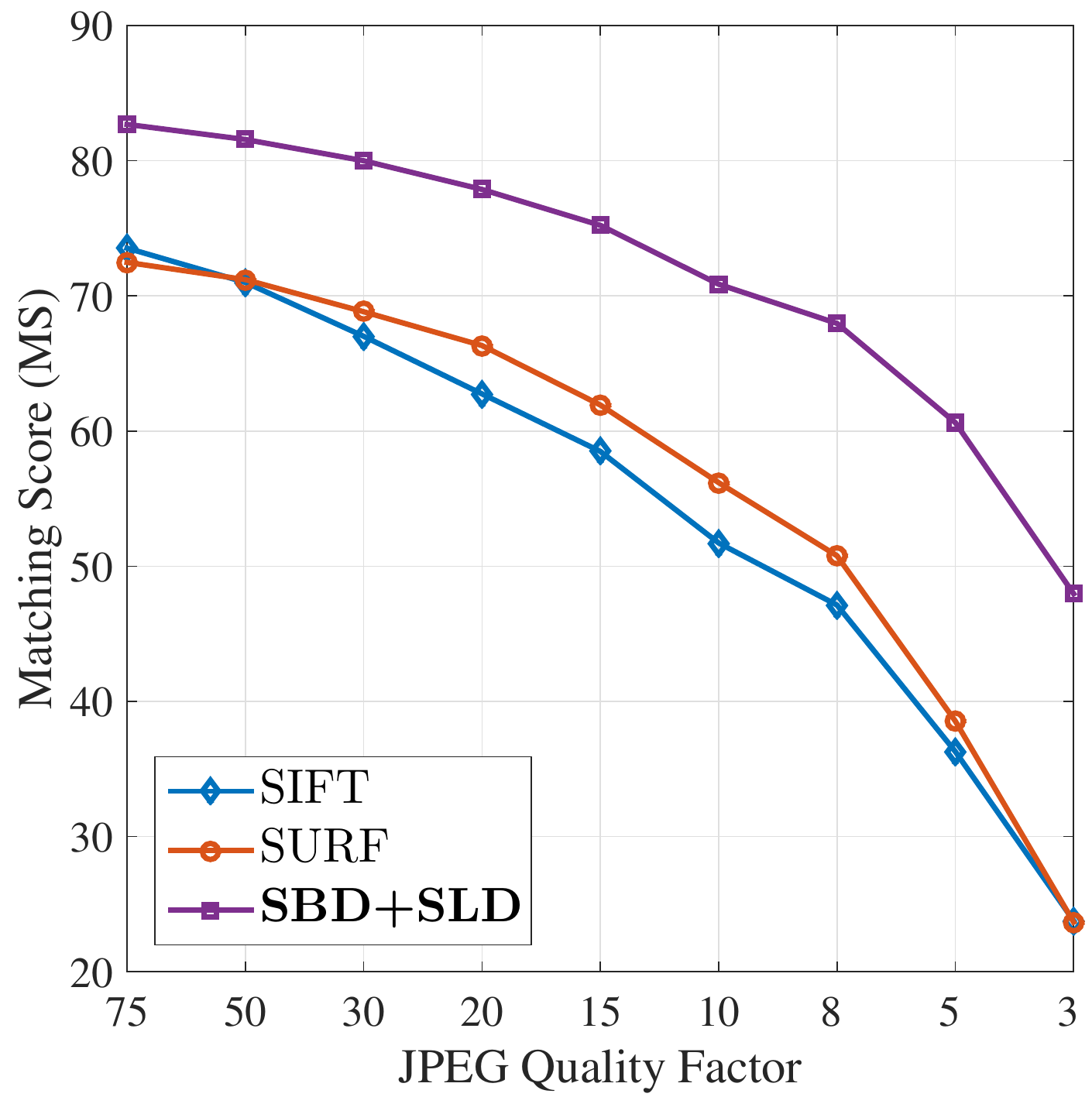}
    \includegraphics[height=0.45\textwidth]{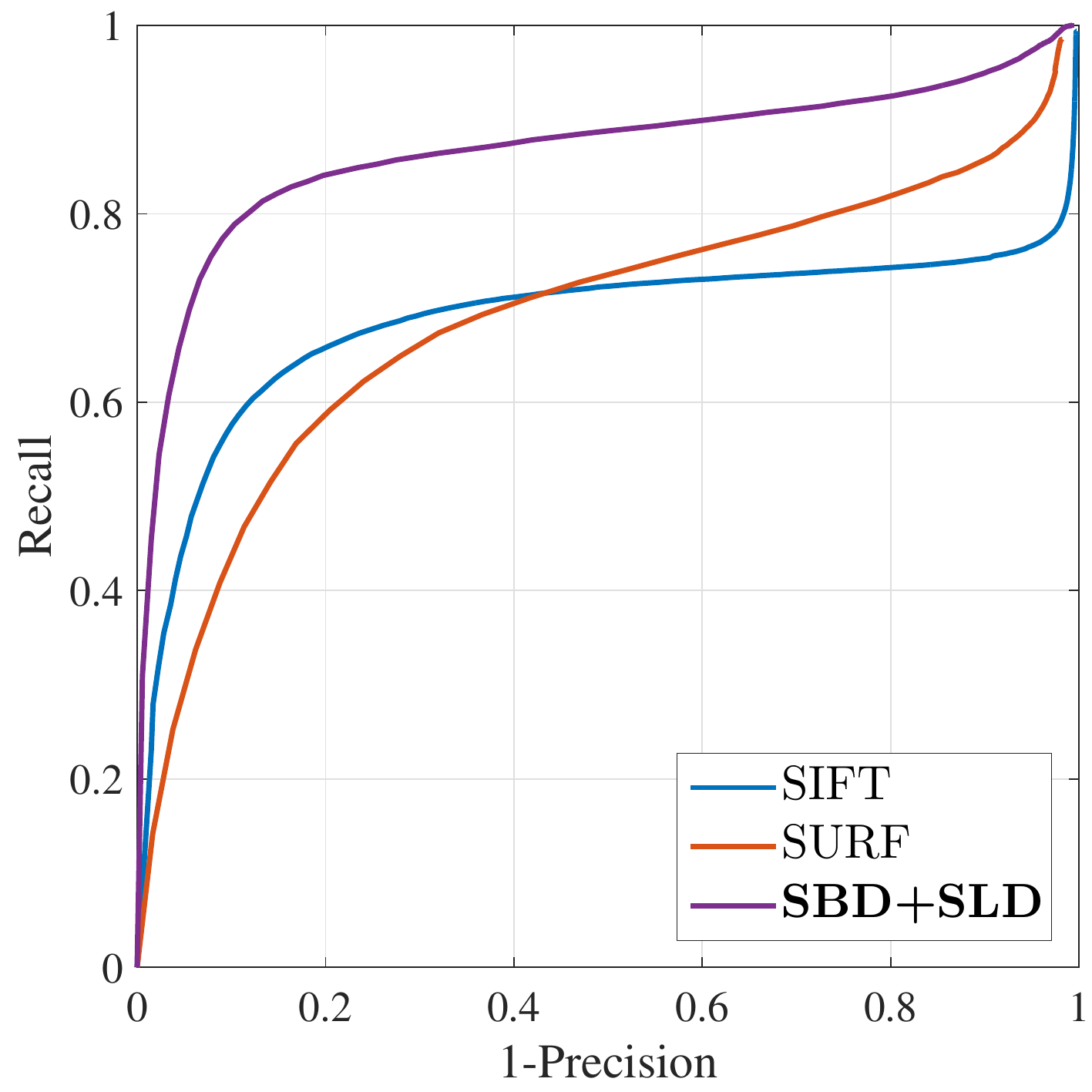}}\\
    \subfloat[Gaussian noise corruption]{
    \includegraphics[height=0.45\textwidth]{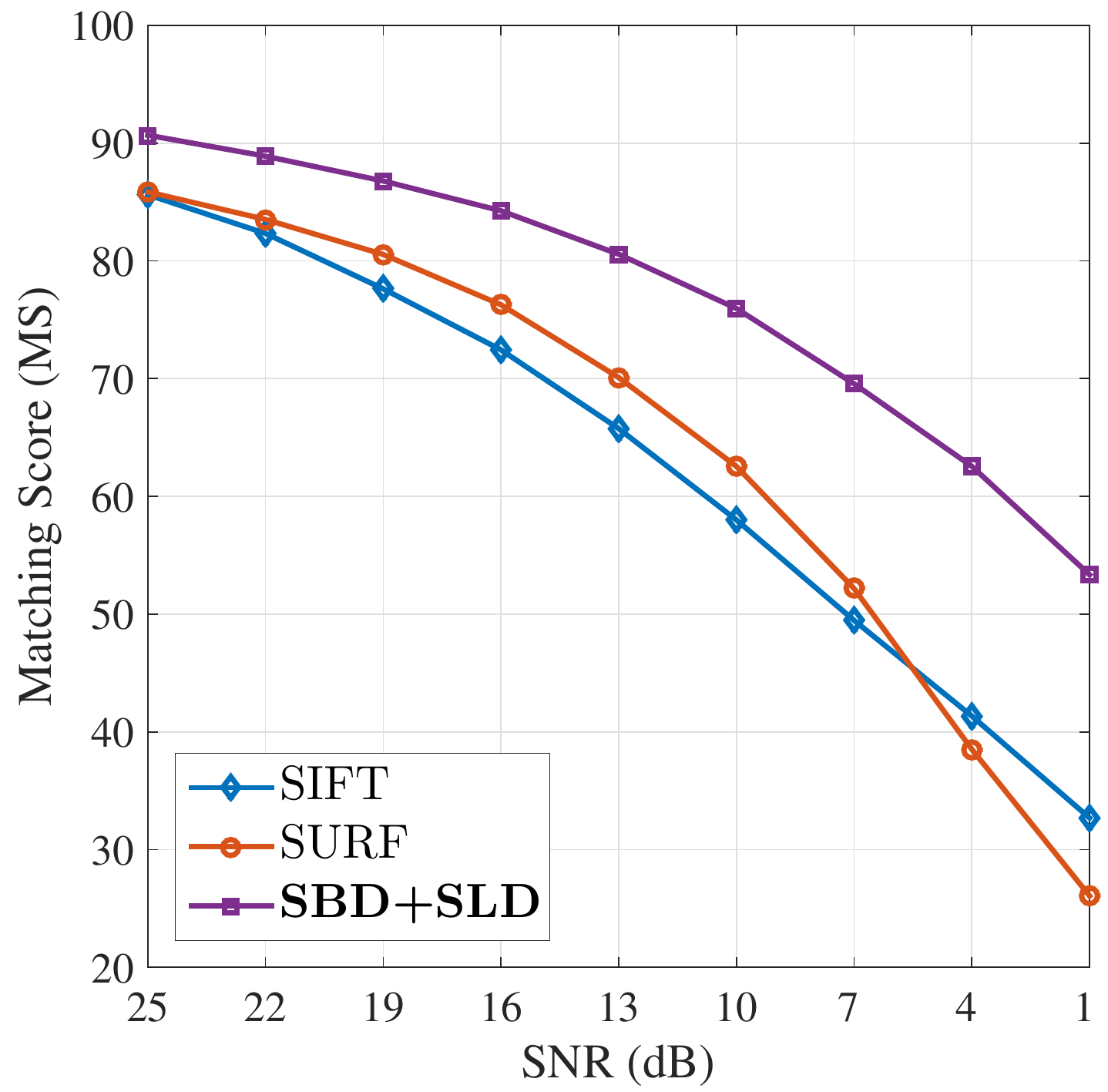}
    \includegraphics[height=0.45\textwidth]{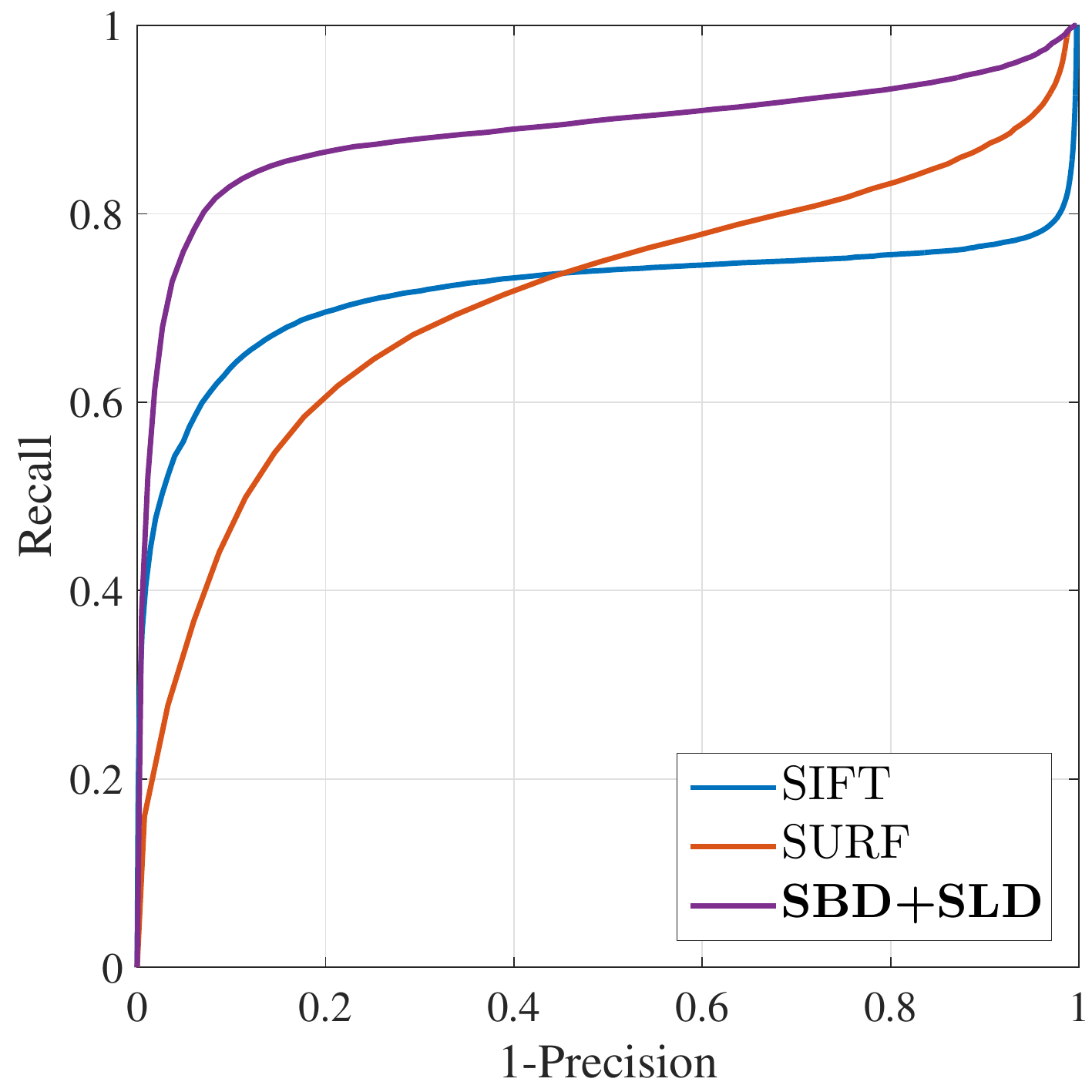}}
    \caption{Comparison of blob detectors with their respective descriptor on the {\it INRIA Copydays dataset}. Left: matching score against amount of compression (a) and noise corruption (b). Right: recall vs 1-precision curve between untransformed and 15 QF compressed (a) and 13 dB of SNR noise corrupted images.}
\label{fig:exp_copydays}
\end{figure}

\section{Experiments on compressed and noisy images}
\label{sec:bdd_compress_image_exp}

In this section we discuss the use of our full pipeline (SBD+SLD) on a larger set of images, and consider the problem of matching images characterised by different levels of compression and noise. We evaluate our results on the \emph{INRIA Copydays} dataset\footnote{The datasets is available at \url{https://lear.inrialpes.fr/~jegou/data.php}}, which contains 157 natural images that are progressively compressed, from 3 (very low quality) to 75 (typical web quality) quality factor (QF). For the evaluation in noisy  environments, the  images were also progressively corrupted with Gaussian noise.

Our method is compared with the  SIFT and SURF methods (DoG+SIFT and fastHessian+SURF, respectively). For the evaluation, we considered the \emph{Matching Score} (MS) \cite{mikolajczyk2005comparison} which is the ratio between the number of correct matches and the number of detected features.
For a visual impression of the overall performances, average values are reported. Fig. \ref{fig:exp_copydays} (left) shows the comparison, where the matching superiority of our approach can be appreciated in both JPEG compression (a) and noise corruption (b).

As a further evidence, we also provide the \emph{recall} vs \emph{1-precision} curve (see Fig. \ref{fig:exp_copydays}, right) obtained when matching the (untransformed) images with the compressed instances (15 quality factor) in (a) and noisy instances (13 dB of Signal to Noise Ratio) in (b). Note that our approach consistently outperforms the competitors.

The results we obtained are in good agreement with the theoretical intuition   that shearlets are an appropriate choice in particular when dealing with noisy and compressed signals.

\section{Conclusions}
\label{sec:bdd-shear_discussion}

In this paper we considered the shearlet representation as a multi-scale framework for the detection and the description of scale-invariant interest points. We first provided a comparative analysis of scale invariance in the scale-space and shearlets domains, in the continuous case --- following the reasoning proposed by Lindeberg \cite{lindeberg2015image}.  Then, we considered a discrete setting and  we addressed the problem of detecting and describing blob-like features by means of the shearlet transform, exploiting it capability of embedding naturally  both scale and orientation information.  More specifically, we
proposed a Shearlet Blob Detector (SBD) algorithm
and a Shearlet Local Descriptor (SLD) algorithm, which we
experimentally assessed by a thorough evaluation on a benchmark dataset. Our algorithm compared favorably with the state of the art, showing a very good tolerance to blur, illumination variations and compression in particular.
We also considered a larger dataset of images affected by different degrees of noise or compression degradation, showing how our shearlet-based pipeline, which includes both detection and description, provided superior results to SIFT and SURF. 

{ In future works different shearlet transform alternatives will be worth investigating. In particular, compactly supported shearlets in the space domain \cite{kittipoom2012construction} have been recently shown to have nice properties for edge detection \cite{kutyniok2014classification} since they could allow us to capture effectively the spatial locality of image features. Since our proposed methods follows exactly the theoretical conceptual path of shearlets, another future work will be to provide a more efficient implementation of our methods based on approximation strategies and optimizations as in \cite{lowe2004distinctive,bay2008speeded}. Moreover, the process of computing the shearlet transform has an intrinsically parallel nature, thus a GPU implementation can provide a dramatic improvement, as shown in \cite{gibert2014discrete} for image denoising. By using the 3D shearlet transform \cite{dahlke2013shearlet}, we also have the interest of extending the proposed shearlet detectors to 3D signals and video image sequences (2D + time).}


\begin{appendices}
\section{Proof of Theorem \ref{th:one}}

\begin{proof}
Let $\mathcal{SH}(f)(a,s,t_1,t_2)$ be the continuous shearlet
transform of $f$ given by~\eqref{eq:cst}.
With the change of variables $\omega_1 = \xi$ and $\omega_2 = v \xi$ whose Jacobian is
\begin{equation*}
\left| \frac{\partial(\omega_1, \omega_2)}{\partial(\xi, v)} \right| = \left| \operatorname{det} \left[ \begin{array}{cc} 1 & 0 \\ v & \xi \end{array} \right] \right| = |\xi|
\end{equation*}
Eq.  \eqref{eq:cst} can be rewritten as
\begin{multline*}
\mathcal{SH}(f)(a,s,t_1,t_2)= a^{3/4}\\ \times \int_{\mathbb{R}} \left\{ \overline{\hat{\psi_2} \left( \frac{v-s}{\sqrt{a}} \right)} \left[ \int_{\hat{\mathbb{R}}} \hat{f} (\xi, v \xi) \overline{\hat{\psi}_1 (a \xi)}  e^{2 \pi i \xi \left( t_1 + v t_2\right)}  |\xi| d\xi \right] \right\}dv
\end{multline*}
where the inner integral
\begin{equation*}
I(a,s,t_1,t_2,v)=  \int_{\hat{\mathbb{R}}} \hat{f} (\xi, v \xi)\overline{\hat{\psi}_1 (a \xi)}  e^{2 \pi i \xi \left( t_1 + vt_2\right)}  |\xi| d\xi
\end{equation*}
is independent on $s$. Now, by recalling the definition of $B[f]$
given by~\eqref{eq:cont_blobness} and
by interchanging the integrals over $v$ and $s$ we obtain,
\begin{multline}
\label{eq:srit}
B[f](a,z_1, z_2) = a^{-1/2}\int_{\hat{\mathbb{R}}}\left\{ I(a,s,az_1,az_2,v) \int_{\mathbb{R}} \overline{\hat{\psi}_2 \left(\frac{v-s}{\sqrt{a}} \right)} ds \right\} dv.
\end{multline}
The change of variable $w= \frac{v-s}{\sqrt{a}}$ gives
\begin{equation}
\label{eq:changevarz}
\int_{\mathbb{R}} \overline{\hat{\psi}_2 \left( \frac{v-s}{\sqrt{a}}\right)} ds = \sqrt{a} \int_{\mathbb{R}} \overline{\hat{\psi}_2(w)} dw =\sqrt{a} \overline{\psi_2 (0)}\ne 0,
\end{equation}
since $\hat{\psi}$ is a bump function.
Next, by plugging Eq.  \eqref{eq:changevarz} into Eq.  \eqref{eq:srit} we obtain
\begin{multline}\label{eq:blob_fourier}
B[f](a,z_1, z_2) = \overline{\psi_2 (0)}\int_{\mathbb{R}} \left[\int_{\hat{\mathbb{R}}}  \hat{f} (\xi, v\xi)\overline{\hat{\psi}_1(a\xi)} e^{2 \pi i a\xi(z_1 + v z_2)}  |\xi|d\xi\right]dv.
\end{multline}
Given$\alpha>0$, let now $f_\alpha$ be  the isotropic dilation of $f$, then
Since  $\hat{f}_{\alpha} (\omega_1, \omega_2) = \alpha^2 \hat{f}(\alpha
\omega_1, \alpha \omega_2)$, we obtain
\begin{multline}
\label{eq:sirtf}
B[f_{\alpha}](a,z_1, z_2) = \overline{\psi_2 (0)}
\int_{\mathbb{R}} \left[ \int_{\hat{\mathbb{R}}} \alpha^2 \hat{f} (\alpha
\xi, v\alpha \xi ) \hat{\psi}_1(a \xi) e^{2 \pi i a \xi (z_1 + v z_2)}
|\xi| d\xi\right]dv.
\end{multline}
Next, the change of variable $\xi^\prime = \alpha \xi$ in the inner
integral gives
\begin{align*}
B[f_{\alpha}](a,z_1, z_2) &=  \overline{\psi_2 (0)}\\ &
\int_{\mathbb{R}} \left[
\int_{\hat{\mathbb{R}}}  \alpha \hat{f}(\xi', v\xi' ) \hat{\psi}_1(a
                                    \alpha^{-1} \xi')  e^{2 \pi i \xi'
                                    \alpha^{-1} a(z_1 + v z_2)}
                                    |\alpha^{-1} \xi'| d \xi' \right] dv\\
& =B[f]( \alpha^{-1} a,z_1, z_2).
\end{align*}
If $f$ is given by~\eqref{eq:2d_sin} , its Fourier transform is the sum of
four Dirac delta at $(\pm \alpha,\pm \beta )/2\pi$. Taking into
account that $\psi_1$ is even, from~\eqref{eq:blob_fourier} we get
\[
B[f](a,z)= \overline{\psi_2 (0)}  \overline{\hat{\psi}_1(\frac{a \alpha}{2\pi})} f(az)
\]

\end{proof}
\end{appendices}

\end{document}